\documentclass{article}

     \PassOptionsToPackage{numbers, compress}{natbib}

\usepackage[final]{neurips_2021}
\usepackage{times}
\usepackage{epsfig}
\usepackage{graphicx}
\usepackage{amsmath}
\usepackage{amssymb}
\usepackage{bbding}
\usepackage{pifont}
\usepackage{caption}
\usepackage{subcaption}

\usepackage[utf8]{inputenc} 
\usepackage[T1]{fontenc}    
\usepackage{hyperref}       
\usepackage{url}            
\usepackage{booktabs}       
\usepackage{amsfonts}       
\usepackage{nicefrac}       
\usepackage{microtype}      
\usepackage{xcolor}         
\usepackage{caption}

\usepackage{wrapfig}

\definecolor{myorange}{RGB}{255,69,0}

\newcommand{\addtext}[1]{\textcolor{black}{#1}}

\renewcommand{\paragraph}[1]{\smallskip \noindent \textbf{#1}}

\makeatletter
\DeclareRobustCommand\onedot{\futurelet\@let@token\@onedot}
\def\@onedot{\ifx\@let@token.\else.\null\fi\xspace}

\title{Look at What I’m Doing: Self-Supervised Spatial Grounding of Narrations in Instructional Videos}

\author{Reuben Tan$^{1}$ \ \ \ \ Bryan A. Plummer$^{1}$ \ \ \ \ Kate Saenko$^{1,2}$ \ \ \ \ Hailin Jin$^{3}$ \ \ \ \ Bryan Russell$^{3}$ \\
$^{1}$Boston University, $^{2}$MIT-IBM Watson AI Lab, IBM Research, $^{3}$ Adobe Research \\
{\tt \small \{rxtan, bplum, saenko\}@bu.edu}, {\tt \small \{hljin, brussell\}@adobe.com} \\
https://cs-people.bu.edu/rxtan/projects/grounding\textunderscore narrations} 

\begin{document}

\maketitle

\begin{abstract}
  We introduce the task of spatially localizing narrated interactions in videos. Key to our approach is the ability to learn to spatially localize interactions with self-supervision on a large corpus of videos with accompanying transcribed narrations. 
To achieve this goal, we propose a multilayer cross-modal attention network that enables effective optimization of a contrastive loss during training. We introduce a divided strategy that alternates between computing inter- and intra-modal attention across the visual and natural language modalities, which allows effective training via directly contrasting the two modalities' representations. We demonstrate the effectiveness of our approach by self-training on the HowTo100M instructional video dataset and evaluating on a newly collected dataset of localized described interactions in the YouCook2 dataset. We show that our approach outperforms alternative baselines, including shallow co-attention and full cross-modal attention. 
\addtext{We also apply our approach to grounding phrases in images with weak supervision on Flickr30K and show that stacking multiple attention layers is effective and, when combined with a word-to-region loss, achieves state of the art on recall-at-one and pointing hand accuracies.
}
\end{abstract}

\section{Introduction}
Content creators often add a voice-over narration to their videos to point out important moments and guide the viewer \addtext{on} where to look. 
While the timing of the narration in the audio track in the video gives the viewer a rough idea of when the described moment occurs, there is not explicit information on where to look. 
We, as viewers, naturally infer this information from the narration and often direct our attention to the spatial location of the described moment. 

\begin{figure}
    \centering
    \includegraphics[width=0.9\linewidth]{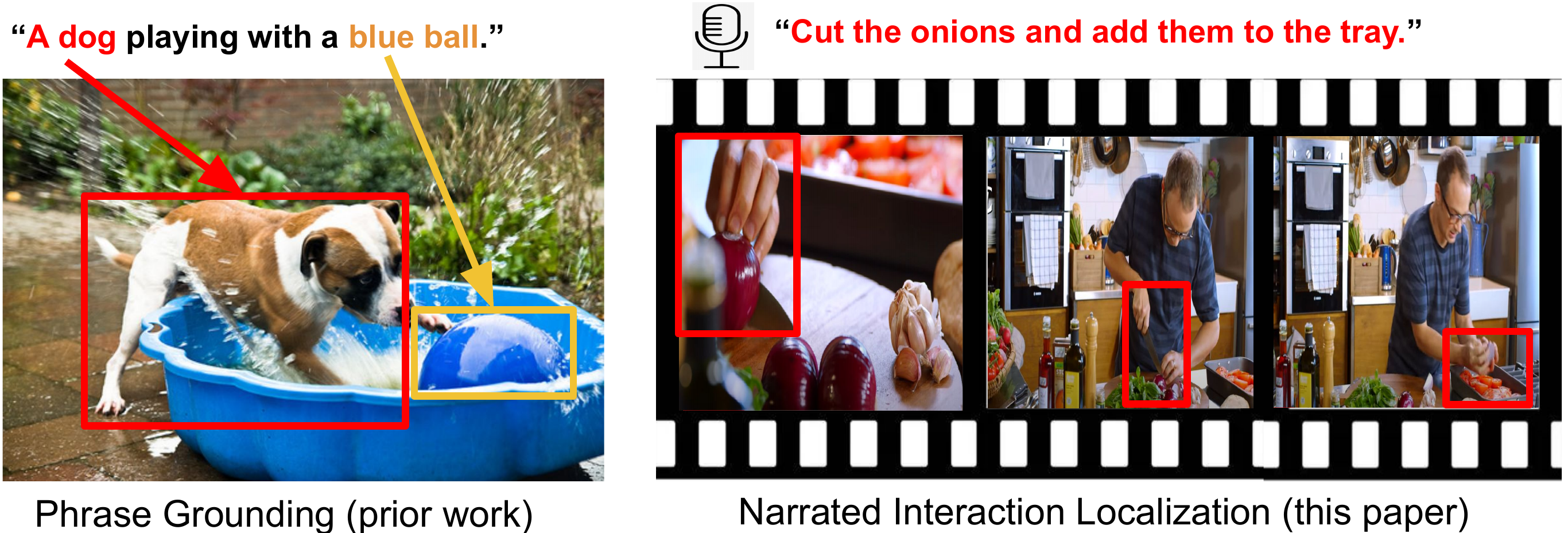}
    \caption{Prior work on \textit{phrase grounding}  focuses on localizing single objects in static images (left, example from Flickr30K~\cite{plummer2015flickr30k}). In this paper, we propose a new task of spatially \textit{localizing narrated interactions} in video (right). In the task of phrase grounding, the goal is to localize nouns or adjective-noun phrases in images. In contrast,
    interaction grounding may involve multiple objects and actions and use entire sentences. Our approach self-trains on a large corpus of narrated videos and does not require supervision. (Video credit: Woolworths \cite{woolworths_vid}) 
    }
    \label{fig:motiv}
    \vspace{-12pt}
\end{figure}

Inspired by this capability, we seek to have a recognition system learn to spatially localize narrated moments without strong supervision.
We see a large opportunity for self-supervision as there is an abundance of online narrated video content. 
In this work, we primarily focus on spatially localizing transcribed narrated interactions in a video, illustrated in Figure~\ref{fig:motiv} (right). Unlike prior phrase grounding work (Figure~\ref{fig:motiv} left), which primarily focuses on matching a noun phrase to an object, our task involves matching entire sentences to regions containing multiple objects and actions. 


Not only is this task integral to advancing machine perception of our world where information often comes in different modalities, it also has important implications in fundamental vision-and-language research such as robotics, visual question answering, video captioning, and retrieval. Our task is challenging as we do not know the correspondence between \addtext{spatial} regions in a video and words in the transcribed narration. 
Moreover, there is a loose temporal alignment of the narration with the described moment and not all words refer to spatial regions (and vice versa). 
Finally, narrated interactions may be complex with long-range dependencies and multiple actions. In Figure~\ref{fig:motiv} (right), notice how ``them'' refers to the onions and that there are two actions -- ``cut'' and ``add''. 

Contrastive learning~\cite{oord2019representation} and computing contextual features via an attention model~\cite{vaswani2017attention} are natural tools for addressing the aforementioned alignment and long-range cross-modal dependency challenges. 
While contrastive learning has been successfully applied for temporal alignment of a video clip with a narration~\cite{miech2020end}, the alignment of \addtext{spatial} regions with \addtext{narrations} has not been addressed. 
Naively applying a joint attention model over the set of features for the two modalities, followed by optimizing a contrastive loss over the two modalities' aggregated contextual features, results in a fundamental difficulty. 
Recall that an attention computation comprises two main steps: (i) selecting a set of features given a query feature, and (ii) aggregating the selected features via weighted averaging.
By selecting and aggregating features over the two modalities during the attention computation and subsequently optimizing a contrastive loss, the model in theory may select and aggregate features from only one modality to trivially align the contextual representations.

\addtext{While there are approaches for grounding natural language phrases in images without spatial supervision, the best-performing approaches involve matching individual words to regions~\cite{gupta2020contrastive} or not using any paired text-image supervision at all~\cite{wang2019without}. This strategy is effective for simpler short phrases, as illustrated in Figure~\ref{fig:motiv} (left). However, consider the interaction ``cut the onions and add them to the tray'', shown in Figure~\ref{fig:motiv} (right). This interaction is described with more complex, compositional language. In the first two frames, there are multiple ``onion" objects visible and the correct one depends on the action being applied. Unlike the task of object grounding, it is not sufficient to simply co-localize all of the mentioned objects. We will show that it is difficult to effectively employ a strategy that matches individual words to regions to localize such complex interactions.}

To address these challenges, we make the following contributions. 
First, we learn to localize the spatial location of a narrated interaction from abundantly available narrated instructional videos~\cite{miech2019howto100m}. Our approach does not require manually collecting sentence descriptions or the locations of the described interactions for training. 
Here, the automatically transcribed narrations allow for self-supervised learning via alignment with the video clip. 

Second, we propose a new approach for directly contrasting \addtext{aggregated} representations from the two modalities while computing joint attention over \addtext{spatial} regions in a video and words in a narration. 
\addtext{We show that optimizing a loss over the aggregated, sentence-level representation allows for a global alignment of the described interaction with the video clip, and offers an improvement over optimizing a matching loss over words and regions.} 
To overcome the network learning a trivial solution while directly contrasting jointly attended features from the two modalities during training, 
we design network attention layers that do not allow feature aggregation across the two modalities. 
Our strategy involves alternating network layers that compute inter- and intra-modal attention. 
Our inter-modal attention layer allows features from one modality to select and aggregate features from only the other modality, and not within its own modality. 
Our intra-modal attention layer selects and aggregates features from within the same modality. This strategy ensures that the output representations do not aggregate features across the two modalities. In combination with stacking the inter- and intra-modal attention layers, our approach attends jointly and deeply and directly contrasts  the resulting contextual representations from the two modalities. 

Finally, we introduce an evaluation dataset that provides bounding box annotations for interactions described by natural language sentences. Our dataset is built on the validation split of the YouCook2 dataset~\cite{youcook2zhou2018} and contains approximately 1000 segments of varying durations.
We demonstrate our approach on our collected evaluation dataset of localized interactions and on localizing objects via the YouCook2-BB benchmark where we show that we outperform shallow and full cross-modal attention. 
\addtext{We also apply our approach to grounding phrases in images with weak supervision on Flickr30K and show that stacking multiple attention layers is effective with our loss. Furthermore, we show that our loss is complementary with the word-to-region loss of Gupta \emph{et al}.~\cite{gupta2020contrastive}, and when combined with it, achieves state of the art on recall-at-one and pointing-hand accuracies.}
\section{Related Work}
\paragraph{Self-supervised learning.} There has been significant progress in natural language processing, resulting in effective and robust learned word representations~\cite{mikolov2013efficient, devlin2018bert, yang2019xlnet, peters2018deep} that achieve state-of-the-art performance on downstream tasks. 
Recently, it has also  garnered a lot of interest in the computer vision community, achieving state-of-the-art performance in unsupervised pre-training of deep visual models. \cite{chen2020simple, he2020momentum, chen2020improved} have demonstrated that effective image representations can be learnt simply by contrasting between augmented views of the same image without labels. \cite{zhao2020distilling,selvaraju2020casting} found that the resulting representations are biased by background pixels and tried to improve foreground object localization via data augmentation or saliency.  In video, self-supervised tasks like pace prediction ad future prediction~\cite{wang2020selfsupvideo, han2020memory, fernando2017self, jabri2020space, benaim2020speednet, wei2018learning} were used for pre-training, while \cite{afouras2020selfsupervised} used self-supervision on audio-visual data for tasks like unsupervised speaker localization. In this work, we propose a self-supervised method to spatially localize activity descriptions in video.
 
\paragraph{Object and action localization.} 
Approaches generally leverage a region proposal network (RPN) \cite{uijlings2013selective, girshick2014rich, girshick2015fast} as well as region-based convolutional neural network (CNNs) to detect objects or localize actions temporally~\cite{xu2017r} or spatio-temporally~\cite{kalogeiton2017action}. However, they are often trained on curated datasets
and are, consequently, limited by a fixed predefined number of object categories in these datasets. \cite{youcook2zhou2018} collected a dataset of cooking videos labeled with descriptions and temporal segments but did not address spatial grounding of activities. Motivated by this limitation, our proposed approach aims to learn to recognize human interactions with objects that belong to the long-tailed distribution from uncurated and unlabeled online videos. 
There have been video datasets collected with bounding box annotations with associated natural language descriptions~\cite{zhou2019grounded}. These datasets provide bounding boxes for noun phrases~\cite{Damen2020Collection,zhou2019grounded} or for egocentric videos~\cite{Damen2020Collection}. 
We collect the first dataset of captioned interactions annotated with spatial bounding boxes for evaluating narrated interactions. 

\paragraph{Vision-and-language.} 
Semantic information from natural language has often been exploited to provide an additional source of supervision for learning visual representations. Such approaches have generally leveraged image~\cite{krishna2017visual,lin2014microsoft,plummer2015flickr30k} and video~\cite{anne2017localizing} datasets that are annotated with natural language descriptions to learn a joint embedding space, where visual and language representations of semantically-similar pairs are close.
Prior approaches have mainly focused on retrieving short clips from a large video corpus \cite{miech2020end} or localizing relevant segments within untrimmed videos \cite{anne2017localizing, chen2018temporally, chen2019weakly}.  However, they do not identify the relevant spatial locations of the described interactions. Some existing works aim to spatially localize a natural language query~\cite{Sadhu_2020_CVPR,zhang2020does,li2017tracking} but focus on objects and train on ground truth bounding boxes, whereas our approach is unsupervised.
Object grounding with weak or no supervision is addressed in~\cite{chen2019weakly,amrani2020self,bertasius2020cobe}. 

To alleviate the costly annotations of curated datasets, \cite{miech2020end} have proposed to learn robust video representations as well as an effective joint video-text embedding space from a large corpus of unlabeled and uncurated instructional videos \cite{miech2019howto100m}. 
In concurrent work, CLIP~\cite{CLIPRadford2021} trained image-text representations on a dataset of 400 million (image, text) pairs collected from the internet. These efforts to learn from ``free'' data are similar in spirit to our work, however, we focus on the localization task.
\section{Self-Supervised Grounding of Narrated Video Interactions} \label{approach}
\begin{figure}[t]
     \centering
         \includegraphics[width=0.9\linewidth]{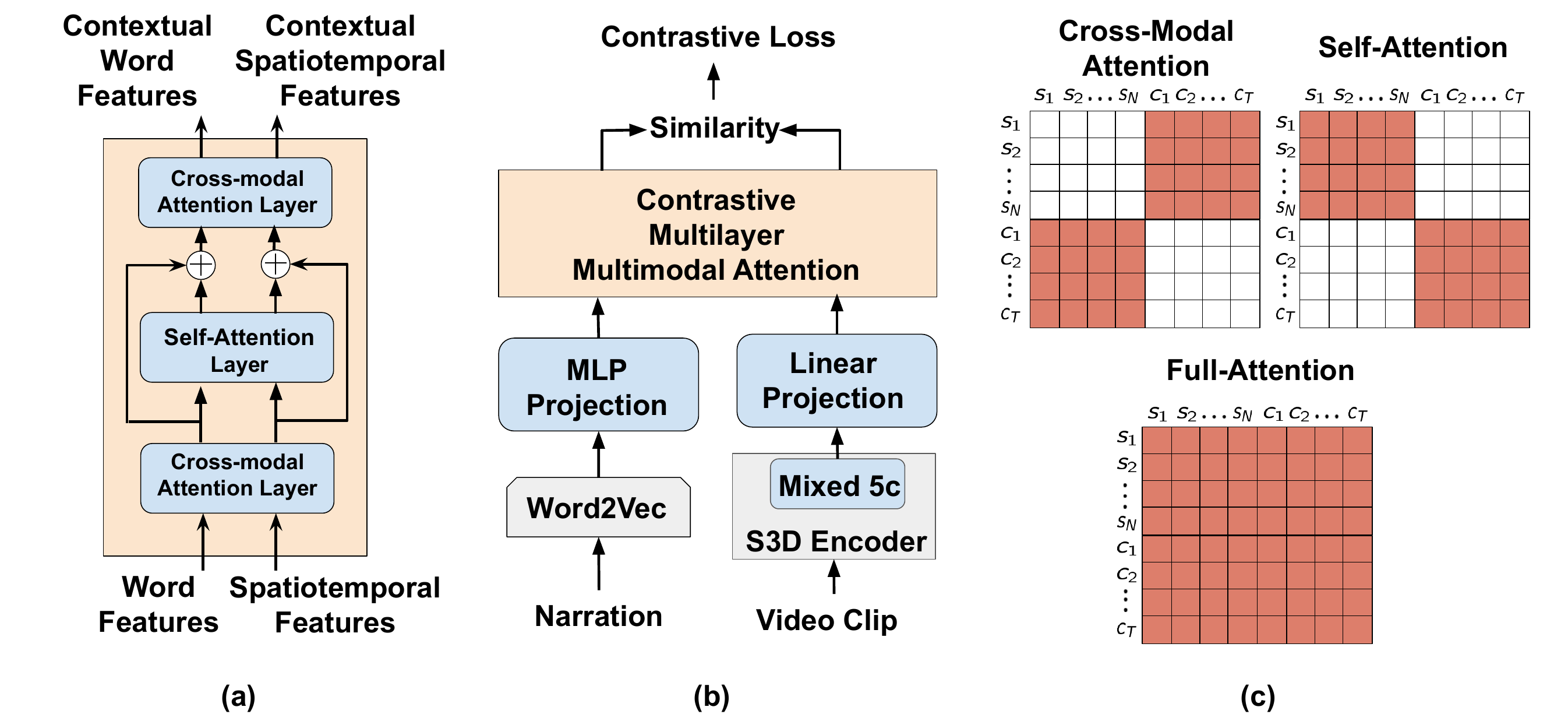}
     \caption{{\bf Approach overview.} 
     (a) Our proposed \textbf{Contrastive Multilayer Multimodal Attention (CoMMA)} module. 
     (b) Our proposed architecture for computing joint attention over narrations and videos. 
     (c) Binary masks for computing the different types of cross-modal attention. The shaded regions represent ``true'' Boolean values where a query feature may select a key.  If the mask value is ``false'', then that corresponding (query, key) pair is not considered in the computation of attention and weighted feature aggregation. 
     }
     \label{model_fig}
     \vspace{-14pt}
\end{figure}

Given a video clip and a narration, our objective is to learn to localize the relevant spatial regions in the clip that correspond to the interaction described in the narration without localization supervision, \emph{e.g}., bounding boxes or segmentation masks. To this end, we propose a novel Contrastive Multilayer Multimodal Attention (CoMMA) module that allows for attention interactions between the features for spatiotemporal regions and words, as well as an effective contrasting of the two modalities' signals. We define two modalities to have attention interactions when their features attend to each other to compute their final contextualized representations. This setup is in contrast to when the two modalities' features only interact when the training loss is computed \cite{miech2020end}. 

Our CoMMA module comprises alternating bidirectional cross-modal attention and self-attention layers (Figure \ref{model_fig}(a)). 
The intuition underlying our proposed approach is that a hierarchical multimodal attention model with multiple stacked layers will encourage a fine-grained alignment between video spatiotemporal regions and narration words and ignore irrelevant regions/words. More specifically, each bidirectional cross-modal attention layer computes new representations for a target modality with the latent representations from the source modality to learn the relevance of each spatiotemporal region to each word or phrase, and vice versa. Additionally, the self-attention layer serves to identify relevant words and regions by aggregating contextual information between the augmented unimodal features. Through repeated attention interactions of the two modalities and aggregation of unimodal contextual information, our CoMMA module becomes more discriminative of irrelevant regions and words. Moreover, our module allows for applying a contrastive loss to the final representations for the two modalities since there is no cross-modal feature aggregation, \emph{i.e}., a modality may self-select features or select features from another modality by attending to them, but it may never select features from both modalities. Our approach goes beyond shallow attention architectures \cite{akbari2019multi} and applying a contrastive loss to independent fixed-length features for each modality (\emph{i.e}., no attention interaction between the two modalities) \cite{miech2020end}. Our multimodal attention module can be easily stacked on top of base feature encoders for our task (Figure \ref{model_fig}(b) shows our full network). Next, we describe our CoMMA module in detail, how we employ a contrastive loss for training, and how we employ our network for the localization task.

\subsection{Contrastive Multilayer Multimodal Attention} \label{localization training}
We formulate each attention layer in our CoMMA module as a key/query/value attention mechanism that is commonly used in attention models \cite{vaswani2017attention}. 
Our module accepts a concatenated sequence of narration word and video spatiotemporal region features and a set of binary attention masks. Let $c_{r}$ be a spatiotemporal feature for region $r$ and $s_j$ be a word feature for the $j$-th word in a narration with $N$ words. We obtain features $c_{r}$ and $s_j$ via base feature encoders that are projected to a common embedding space using linear and MLP layers. We form matrix inputs $C_0$ and $S_0$ by stacking the spatiotemporal and word features $c_{r}$ and $s_j$, respectively, as column vectors.

In Figure \ref{model_fig}(c), the shaded regions of the binary masks indicate which features interact during the attention computation for a layer. Specifically, the cross-modal attention (CA) mask $M^{CA}$ computes the attention over the words with respect to each region and vice versa. Note that this setup differs from the full-attention (FA) mask $M^{FA}$ (illustrated in Figure \ref{model_fig}(c) bottom) that computes attention over the \emph{entire} multimodal sequence of words and regions and the self-attention (SA) mask $M^{SA}$ (illustrated in Figure~\ref{model_fig}(c) top-right) that computes attention within each modality. We extend the mechanism of Vaswani \emph{et al}.~\cite{vaswani2017attention} by incorporating a mask that modulates the attention computation. Formally, let $K$, $Q$, $V$ be the matrices of features for the keys, queries and values, respectively (the individual features are stacked as column vectors). Let the masked attention mechanism be represented as:
\begin{equation} \label{attn_func}
    Attn(K, Q, V, M) =V \operatorname{softmax}\left(\frac{\left(Q^\top K\right) \odot M}{\sqrt{D}}\right)
\end{equation} 
where $M$ is any one of the binary masks shown in Figure \ref{model_fig}(c), $D$ denotes the scalar dimensionality of the query features, and $\odot$ is the Hadamard product.

For simplicity, let matrix $Y^{SA}_0 = [C_0, S_0]$ denote the stacked input features from both modalities. In addition, let $W_{K,l}^{i}$, $W_{Q,l}^{i}$ and $W_{V,l}^{i}$ denote the projection matrices for keys, queries, and values, respectively, for attention type $i$ and layer $l$. Here, $i$ can be used to represent a cross-attention (CA) or self-attention (SA) layer and $l$ is the layer number. We compute the output of the cross-modal attention layer by passing in the initial input features or contextualized feature outputs of the self-attention layer:
\begin{equation} \label{eq2}
    Y^{CA}_{l+1} = CrossAttn(Y^{SA}_l) = Attn(W_{K,l+1}^{CA} Y^{SA}_l, W_{Q,l+1}^{CA} Y^{SA}_l, W_{V,l+1}^{CA} Y^{SA}_l, M^{CA})
\end{equation}
where $M^{CA}$ is the cross-modal attention mask. Note that if the input to Equation (\ref{eq2}) is $Y_l^{SA} = [C_l, S_l]$, then the output will be $Y_{l+1}^{CA} = [S_{l+1}, C_{l+1}]$, \emph{i.e}., (contextualized) video features become (contextualized) language features and vice versa. We emphasize that in the cross-attention layer, there is no mixing of features from different modalities where an elementwise sum is applied to features across the different modalities. Consequently, it allows both modalities to attend to each other without leaking any information between them. We include a mathematical formalization of this operation in the supplementary. The similarity scores computed between the queries and the keys as the scaled dot product in Equation (\ref{attn_func}), act as a soft attention mechanism to measure the relevance of each word with respect to a region. The softmax-normalized scores are multiplied with the value (words) vectors to compute a region-specific narration representation. To illustrate this concept, consider a YouTube video clip about cooking fried chicken. The words `fried' and `cook' will have a higher similarity score to regions in the video clip than `subscribe' since some regions will be highly relevant to the former.  Conversely, the modalities of the keys and queries can be swapped to compute word-specific clip representations.

To aggregate contextual information over the augmented representations of the two modalities, the outputs of the cross-modal attention layer are passed into a self-attention layer. The self-attention mask $M^{SA}$ ensures that attention is only computed between elements of the same modality. Formally, this computation can be represented as 
\begin{equation}
    Y^{SA}_{l} = SelfAttn(Y^{CA}_{l}) = Attn(W_{K,l}^{SA} Y^{CA}_l, W_{Q,l}^{SA} Y^{CA}_{l}, W_{V,l}^{SA} Y^{CA}_{l}, M^{SA}) + Y^{CA}_{l}
\end{equation}
where $M^{SA}$ is the self-attention mask. Its output is passed into the next cross-attention layer. We denote the final output of our module after $L$ layers as $C_L$ and $S_L$ where $C_{L} \in \mathbb{R}^{D X T}$ is the set of contextualized spatiotemporal region features and $S_L \in \mathbb{R}^{D X N}$ is the set of contextualized word representations. 


\paragraph{Comparison to multimodal attention models.} \label{cross_modal_baselines}
To evaluate the importance of multimodal attention modules for localizing narrations, we briefly compare CoMMA against other state-of-the-art variants. Akbari et al. \cite{akbari2019multi} is an image-level approach that utilizes features from different levels of an encoder. However, unlike CoMMA, it only utilizes a single round of cross-modal interaction. Another highly relevant work is the Contrastive Bidirectional Transformer (CBT) \cite{sun2019learning}. Despite the similarities in attention layers, the CBT model computes self-attention over the entire multimodal sequence (similar to the full-attention mask in Figure~\ref{model_fig}(c)) and aggregates a summary knowledge into a sentinel vector. In contrast, CoMMA allows for repeated attention interactions without mixing of features from different modalities. We show in Section~\ref{quant_results} that this mechanism is critical for localization.
\vspace{-5pt}
\subsection{Contrastive loss and inference}
We train our proposed multimodal attention module by contrasting between positive and negative pairs of video clips and narrations. Formally, we aim to learn language-aligned visual representations for video clips such that features for corresponding videos and narrations are similar and non-corresponding features are dissimilar. $C_{L, r}$ denotes the column vector for the final contextualized feature for region $r$ and $S_{L,j}$ denotes the column vector for the final contextualized representation for word $j$. The final video clip representation $\Hat{C}$ and narration sentence representation $\hat{S}$ are computed by mean-pooling over the spatiotemporal regions \addtext{$\Hat{C} = \frac{1}{R}\sum_{r=1}^R  C_{L, r}$} and words \addtext{$\hat{S} = \frac{1}{N}\sum_{j=1}^{N} S_{L,j}$} respectively. Let $\left(\hat{C}^{(i)},\hat{S}^{(i)}\right)$ be the $i$-th training example pair. We adapt the InfoNCE loss \cite{oord2019representation} by defining the sentence loss $\mathcal{L}_{sent}$ as the sum of log ratios over self-training pairs:
\begin{equation}
\label{eq:video_objective_function}
\mathcal{L}_{sent} =
-\sum_{i=1}^{n} \log \left(\frac{\exp\left(\hat{C}^{(i)}\cdot\hat{S}^{(i)}\right)}{\exp\left(\hat{C}^{(i)}\cdot\hat{S}^{(i)}\right) + \sum\limits_{m\sim \mathcal{N}_{i,S}} \exp\left(\hat{C}^{(i)}\cdot\hat{S}^{(m)}\right) + \sum\limits_{m\sim \mathcal{N}_{i,C}} \exp\left(\hat{C}^{(m)}\cdot\hat{S}^{(i)}\right)} \right)
\end{equation}
where the negative sets $\mathcal{N}_{i, C}$ and $\mathcal{N}_{i,S}$ comprise indices for non-corresponding video clip and narration pairs for the $i$-th training sample, and $n$ denotes the total number of training samples. Note that our CoMMA module is applied to negative pairs as well. \addtext{In contrast to the word-level objective (defined in Section~\ref{quant_results}) used in \cite{gupta2020contrastive} which contrasts word features separately, we empirically find that contrasting aggregated features in the sentence loss $\mathcal{L}_{sent}$ better handles when the temporal alignment between narrations and videos is noisy.} 

\paragraph{Inference.} During inference, given a natural language sentence and a video clip, we aim to localize the salient spatial regions across the frames in the clip. Beginning from the attention weights of the last cross-modal attention layer, we apply attention rollout \cite{abnar2020quantifying} to obtain the final attention heatmap over all spatiotemporal regions. In attention rollout, the attention weight matrices from all cross-attention and self-attention layers are multiplied recursively to yield the output localization scores $A$ for each spatiotemporal region, where $A = \prod_{l=0}^L  W_{l}$ for attention weights $W_l$ from the $l$-th layer. The resulting localization scores aggregate the total amount of attention by the entire set of spatiotemporal and word features assigned to a query feature.

\section{Experiments} \label{quant_results}

\addtext{\paragraph{Word-level loss.}}
\addtext{We experiment with a word-level loss that is used in state-of-the-art phrase grounding models~\cite{gupta2020contrastive, akbari2019multi}, which aims to learn an alignment between each word in the narration and all spatial regions. Gupta \emph{et al}.~\cite{gupta2020contrastive} has demonstrated that the word-level loss is effective at localizing noun phrases in images, which seeks to maximize the mutual information between each word and region features for corresponding video clip and narration pairs. } We also consider an extension of the word-level loss where we incorporate the output contextualized representations from our CoMMA module. Let $N$ denote the number of words in the narration, and $n$ the total number of training samples.  $S^{(i)}_{L, j}$ denotes the representation output of our CoMMA for the j-th word of the i-th training sample. Finally, for a given input word $S^{(i)}_{0,j}$, we compute its value representation with a multilayer perceptron (MLP): $\bar{S}^{(i)}_{j} = MLP(S^{(i)}_{0,j})$. Then, the word-level loss is formulated as the sum of log ratios:
\addtext{\begin{equation} \label{eq: word loss}
    \mathcal{L}_{word} = -\sum_{i=1}^{n} \sum_{j=1}^{N} \log{\left(\frac{\exp{\left(S^{(i)}_{L, j}\cdot \bar{S}^{(i)}_{j}\right)}}{\exp{\left(S^{(i)}_{L, j}\cdot \bar{S}^{(i)}_{j}\right)} + \sum\limits_{m \sim \mathcal{N}_i} \exp{\left(S^{(m)}_{L, j}\cdot \bar{S}^{(i)}_{j}\right)}}\right)}
\end{equation}}

\addtext{where the negative set $\mathcal{N}_i$ of the $i$-th training sample comprises indices for the word features that are attended to by non-corresponding video clips. }
\vspace{-5pt}
\addtext{\subsection{Grounding interactions and objects in video}\label{sec:video_experiments}}
\paragraph{Self-training and evaluation datasets.}
We self-train our proposed model on instructional videos from the HowTo100M dataset~\cite{miech2019howto100m}. To reduce the computational burden of training over the entire 100M clips, we identify a set of clips from the dataset that roughly aligns with the YouCook2 dataset. 
To achieve this goal, we extract a list of verb and nouns from the vocabulary of YouCook2 and filter HowTo100M video clips with narrations that contain these words. Our final pre-training set comprises approximately 250,000 video clips. We use the publicly available base video and language feature encoders \cite{miech2020end} that are trained on the entire HowTo100M dataset, which provides a good initialization for our learning task.

\addtext{For evaluation, we introduce a new evaluation dataset, YouCook2-Interactions, that provides spatial bounding box annotations for interactions that are described by a natural language sentence. Our dataset is built on the validation split of the YouCook2 dataset~\cite{youcook2zhou2018}. We describe our dataset in full in the supplemental.} Please see more details of the self-training and our collected dataset in the supplemental.
We evaluate using our collected YouCook2-Interactions dataset for localizing narrated interactions. 

In addition to localizing interactions given a sentence narration, we also consider localizing objects in video given a single object keyword. 
For localizing single objects, we evaluate on the YouCook2-BB dataset~\cite{ZhLoCoBMVC18}. The YouCook2-BB dataset augments the original YouCook2 dataset with bounding box annotations for object locations. In our experiments on YouCook2-BB, we only evaluate our approach on the validation split since the test split annotations are not readily available.

\paragraph{Evaluation criteria.}
For both \addtext{video} tasks, we evaluate the quality of the output detections using the pointing hand accuracy criterion. Specifically, given the ground-truth bounding box, we consider our model to have a ``hit" if the pixel with the highest co-attention similarity score lies within the box. Otherwise, it is a ``miss". The final localization accuracy is computed as the ratio of hits to the total number of hits and misses $\frac{\text{\# hits}}{\text{\# hits} + \text{\# misses}}$.

\paragraph{Implementation details.} We use publicly available implementations for the separable 3D CNN-gated (S3D-G) encoder~\cite{xie2018s3d} and a shallow language encoder \cite{miech2020end} built on top of Word2Vec \cite{mikolov2013efficient} embeddings for our video and language models, respectively. To train our proposed model, we set a learning rate of 1e-4 and optimize the model using the AdamW optimizer \cite{loshchilov2017decoupled} with one-epoch linear warmup. In our experiments, we also explore using two CoMMA modules but, due to the large memory requirements, the batch size has to be reduced significantly. Results from our initial experiments and prior work on contrastive learning show that a large batch size is crucial to achieving strong performance. More details are in the supplemental.

During inference, since the temporal and spatial dimensions of the attention heatmap have been downsampled from the original input resolution, the localization scores are temporally and spatially interpolated back to the input resolution. We return the final spatiotemporal location as the mode over all output scores.

\paragraph{Baselines.} 
In our evaluations, we compare our approach against center prior, optical flow, and state-of-the-art image-ground baselines. In the center prior baseline, we return the center pixel for each frame. 
This is a simple yet competitive baseline that performs well due to the nature of these videos, where the main subject is generally in the center of the camera view. 
As most of the interactions involve motion in cooking videos, we compute an optical flow baseline. For this baseline, we generate an optical flow map for each frame using the OpenCV implementation of the Lucas-Kanade optical flow estimation method \cite{opencv_library,lucas1981iterative}. Finally, we select the pixel with the highest magnitude as the location prediction.

Additionally, we compare against four cross-modal attention baselines. First, we compare against the model of Miech \emph{et al}.~\cite{miech2020end} adapted to our task ("MIL-NCE"). We adapt their publicly available pretrained model by removing the penultimate global average pooling layer in the S3D-G video encoder. Consequently, the output of the video encoder is a set of spatiotemporal features that are projected into the joint embedding space. Then, we compute the dot product between the sentence feature and the spatiotemporal features before performing linear interpolation to determine the mode pixel.
Second, we compare against a baseline that computes full attention and leverages a sentinel vector ("Full attention + sentinel") to predict the similarity of the two modalities' features; an MLP is used to convert the output contextualized sentinel vector into a scalar MI-score which is then passed to an InfoNCE-like loss (inspired by CBT \cite{sun2019cbt}).
Third, we compare our multilayer model against a shallow cross-modal attention layer (``Shallow attention'') inspired by Akbari \emph{et al}.~\cite{akbari2019multi}. To maintain fair comparisons, we implement a similar version of their model except we use for the base features the convolutional feature map outputs of the S3D video encoder. Finally, we include comparisons against a state-of-the-art phrase grounding model \cite{gupta2020contrastive}. We apply the model out of the box to our task with the exception of using our features for fair comparison.

\begin{table}
\centering
 {
\caption{
{\bf Interaction localization evaluation on our YouCook2-Interactions dataset.} Our approach outperforms baselines, including shallow and full attention.
}
\label{table:interaction-localization}
 \begin{tabular}{| c | c | c |} 
 \hline
 Approach & Training Loss & Localization Accuracy (\%)\\
 \hline
 Center Prior & - & 31.81  \\ 
 \hline
 Optical flow & - & 18.75 \\
 \hline
 MIL-NCE \cite{miech2020end} & Sentence level & 27.18\\
 \hline
 \addtext{Contrastive phrase grounding \cite{gupta2020contrastive}} & \addtext{Word level} & \addtext{24.04} \\
 \hline
 Shallow Attention & Sentence level & 48.30 \\
 \hline
 Full attention + sentinel & Sentence level & 39.03 \\
 \hline
 \addtext{CoMMA (Ours)} & \addtext{Word level} & \addtext{34.91}\\
 \hline
 CoMMA (Ours) & Sentence level & \textbf{55.80}\\
 \hline
\end{tabular}
}
\vspace{-10pt}
\end{table}

\begin{table}
\centering
\caption{
{\bf Spatiotemporal self-attention ablation and attention layer ablation.}  
(Left) We report an ablation of the different attention module components in our model. We find that multiple stacked layers outperform shallower architectures. 
(Right) We ablate different types of self-attention over spatiotemporal features and find computing full self-attention over spatiotemporal features works best.
}
\label{table:video_ablation}
\begin{tabular}{| c | c | c| c|} 
 \hline
 Cross- & Cross- & Self- & Localization \\ 
 Modal & Modal & Attention & Accuracy (\%) \\
 Attention 1 & Attention 2 &  & \\ 
 \hline
  \ding{51}  & \ding{55} & \ding{55} & 48.30\\ 
 \hline
 \ding{51} & \ding{51} & \ding{55} & 51.09\\
 \hline
\ding{51} & \ding{51} & \ding{51} & \textbf{55.80}\\ 
 \hline
\end{tabular}
\quad
\begin{tabular}{| c | c |} 
 \hline
 Self-Attention  & Localization \\
 Type &  Accuracy (\%)\\
 \hline
 Spatial & 53.88  \\ 
 \hline
 Temporal & 52.72 \\
 \hline
 Spatial + Temp.\ & 52.15 \\ 
 \hline
 Spatiotemporal & {\bf 55.80} \\
 \hline
\end{tabular}
\vspace{-10pt}
\end{table}

\paragraph{Interaction localization.} We report in Table~\ref{table:interaction-localization} results of our approach and the baselines on the task of interaction localization. Our proposed approach outperforms the center prior baseline by a significant margin of 20\% on the task of localizing interactions given narration sentences. Surprisingly, computing the region with the maximum degree of motion through optical flow does not provide a good proxy for localization. The large performance gain achieved by our proposed approach over the optical flow baseline suggests that tracking regions with maximum movement alone is insufficient for our task. \addtext{In addition, we observe that our proposed approach outperforms a state-of-the-art phrase grounding model~\cite{gupta2020contrastive}, which is trained via optimizing the word-level loss (Equation~\ref{eq: word loss}), by a large margin. One possible reason is that enforcing an alignment between words and regions is not suitable for grounding interactions in videos, especially when there is a weak temporal alignment between clips and narrations. This is corroborated by a huge drop of 20\% in localization accuracy when our CoMMA module is trained with the word-level loss instead.} 

The difference in performance obtained by our model and the MIL-NCE baseline suggests that a model trained for retrieval may not be focusing on the relevant regions that are described by the sentence narration. Our proposed approach outperforms the ``full attention + sentinel'' baseline by 15\%. This result suggests that aggregating contextual information into a sentinel vector causes the model to lose fine-grained information required for localization.  We note that, similar to Sun \emph{et al}.~\cite{sun2019cbt}, this baseline outputs a scalar value that is plugged in directly to the InfoNCE loss; it does not compute a cosine similarity between the video clip and natural language query contextualized features. We hypothesize that computing this MI-like score (instead of directly contrasting the two modalities’ features) may somewhat mitigate this baseline from catastrophic failure where features from both modalities are aggregated during the attention interaction process.

\paragraph{Multilayer attention and spatiotemporal self-attention ablations.} 
We report in Table~\ref{table:video_ablation} (left) an ablation over stacking the different attention layers in our network. Notice that the localization accuracy increases as we stack more attention layers in our network, suggesting that deeper cross-modal attention models help to improve the alignment between the different modalities. We report in Table 2 (right) results from an ablation of the different forms of self-attention over spatiotemporal regions. It is interesting to observe that by applying self-attention to spatial regions alone, the model performs competitively by obtaining a localization accuracy of 53.88\%. We obtain our best model when self-attention is computed over all spatiotemporal regions. This finding suggests that reasoning about temporal context is important but it has to be done in conjunction with spatial context reasoning.

\paragraph{Object localization.} Besides localizing full narrated sentences, we also evaluate our model's capability to localize single objects. Specifically, we compare our approach against the state-of-the-art "not all frames are equal" (NAFAE) model~\cite{shi2019not}. 
We report results in Table~\ref{table:object-localization}. \addtext{Despite being trained with a sentence-level loss, we observe in Table~\ref{table:object-localization} that our approach successfully localizes objects corresponding to single noun words.} Notably, our proposed approach outperforms the NAFAE baseline. We give more details of the NAFAE baseline in the supplemental.

\begin{wraptable}{r}{0.35\textwidth}
\vspace{-14pt}
\centering
    {
\caption{
{\bf Object localization evaluation on the YouCook2-BB dataset.} 
By self-training on narrations, our approach learns to localize objects and outperforms the baseline.
}
\label{table:object-localization}
 \begin{tabular}{| c | c |} 
 \hline
 Approach  & Full Loc.\ (\%)\\
 \hline
 NAFAE~\cite{shi2019not}  & 46.95\\
 \hline
 Ours  & {\bf 59.25}\\ 
 \hline
\end{tabular}
}
\vspace{-20pt}
\end{wraptable}

\begin{figure*}[t]
\begin{center}
\includegraphics[width=\linewidth]{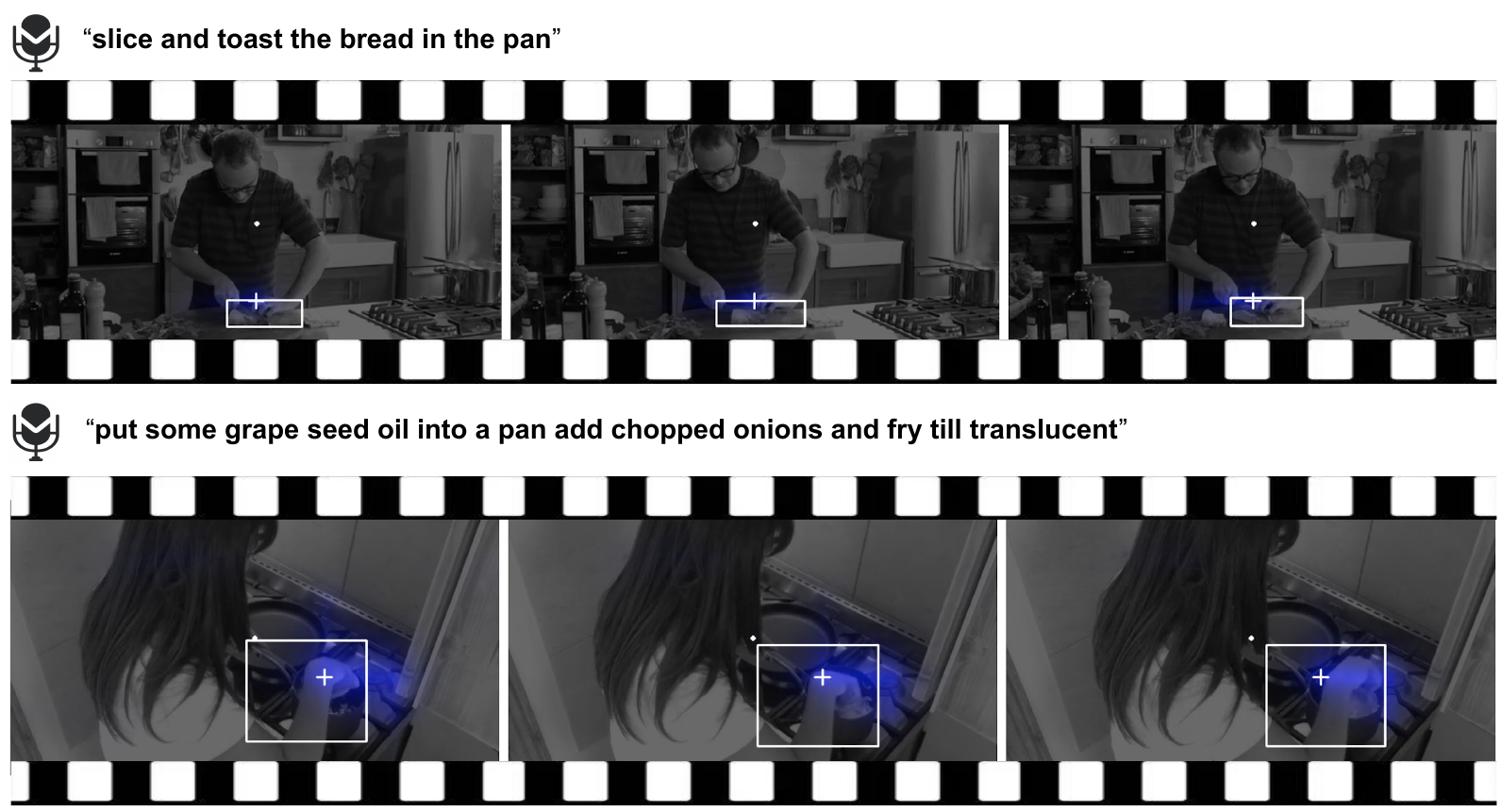}
\end{center}
   \caption{
   Examples of correct localization predictions by our model. Our approach localizes narrated interactions involving a person's hands and interacted objects. The boxes in these visualizations denote the ground-truth bounding boxes for the transcribed narrations. The white circle and cross indicate the locations of the center prior and the mode pixel of the computed attention score map, respectively. Finally, the blue regions indicate the regions that our CoMMA module determines to be most relevant to the transcribed narrations. (Video credit: Woolworths \cite{woolworths_vid} and Alicia Kirby \cite{alicia_vid})
   }
\label{fig:language-maps1}
\vspace{-10pt}
\end{figure*}

\paragraph{Qualitative results.} 
We show qualitative results in Figure \ref{fig:language-maps1}. Each frame is overlaid with the computed attention score map to depict the visual cues that our model uses to make its prediction. The white circle indicates the location of the center prior and the cross denotes the mode pixel of the computed attention score map. Notice how our approach correctly localizes the described interaction. 
\vspace{-3 pt}
\addtext{\subsection{Weakly supervised phrase grounding in images}\label{sec:image_experiments}} 
\addtext{We evaluate our approach on the task of weakly supervised phrase grounding to determine its capability to generalize to localizing noun phrases in images. For this task, we adopt the same setup as in prior work \cite{gupta2020contrastive, datta2019align2ground, fang2015captions} by training on MSCOCO \cite{lin2014microsoft} and evaluating on Flickr30k \cite{plummer2015flickr30k}.} 

\addtext{We report results on the standard recall-at-K (R@K) and pointing hand recall accuracy criteria~\cite{gupta2020contrastive}. For training, we optimize the sum of the word-to-region matching loss and our sentence loss: $\mathcal{L}(\theta) = \mathcal{L}_{word} + \lambda \mathcal{L}_{sent}$. We set $\lambda$ to be 0.005 in our experiments. We compare our approach against state-of-the-art phrase grounding models. Details of how the model in \cite{gupta2020contrastive} is extended with our CoMMA architecture as well as a description of the baselines are included in the supplemental. We note that the value projection layer for region features in \cite{gupta2020contrastive} is omitted to incorporate CoMMA. Finally, we build on the same set of visual and language features from \cite{gupta2020contrastive} for fair comparison.} 

\paragraph{Results and analysis.} We report results in Table~\ref{table:image-phrase-grounding}. As mentioned above, omitting the region value projection layer results in a performance drop, as evidenced by the results obtained by using a single cross-attention layer. Adding a second cross-attention layer generally helps to improve the localization accuracy and leads to notable increases in R@1 and pointing hand recall accuracies, surpassing state-of-the-art \cite{gupta2020contrastive} on those criteria. We note that our full CoMMA model with unimodal self-attention layers hurts performance (R@1 = 48.81\%). One possible reason is that optimizing CoMMA on this task is challenging and it may be overfitting. However, it is very promising that adding more layers works out of the box in general on a task that is typically addressed by shallow attention models. In contrast, we show in the supplemental that stacking multiple layers in our default CoMMA architecture when trained with the sentence-level loss $\mathcal{L}_{sent}$ only results in improved R@K localization accuracies. 

Finally, we also observe that, unlike the observations in the video experiments, the word-level loss is critical for phrase grounding in images.  We hypothesize that this finding is a result of the captions consisting primarily of noun phrases and being strongly correlated with the images. In contrast, the video task requires spatially localizing different interactions that are mentioned in a narration, which are more semantically complex as compared to noun phrases, making the word loss less effective.

\begin{table}
\centering
 {
\caption{
{\bf Weakly supervised phrase grounding experiments on Flickr30K.} Adding additional cross-attention layers improves performance on the R@1 and pointing recall criteria. We include a description of the visual features and baseline approaches in the supplemental. IN and VG denote ImageNet and Visual Genome respectively.
}
\label{table:image-phrase-grounding}
 \begin{tabular}{| c | c | c |  c| c | c|} 
 \hline
 Approach & Visual Features &  R@1 & R@5 & R@10 & Pt Recall\\
  \hline
 \addtext{Fang \emph{et al} \cite{fang2015captions}} & \addtext{VGG-cls (IN)} &  - & - & - & \addtext{29.00} \\
 \hline
 \addtext{Akbari \emph{et al} \cite{akbari2019multi}} & \addtext{VGG-cls (IN)} &  - & - & - & \addtext{61.66} \\
  \hline
 \addtext{Akbari \emph{et al} \cite{akbari2019multi}} &  \addtext{PNAS Net (IN)} &  - & - & - & \addtext{69.19} \\
 \hline
 \addtext{Align2Ground \cite{datta2019align2ground}} & \addtext{Faster-RCNN (VG)} &  - & - & - & \addtext{71.00}\\ 
 \hline
 \addtext{Gupta \emph{et al} \cite{gupta2020contrastive}} & \addtext{Faster-RCNN (VG)} &  \addtext{51.67} &\addtext{\textbf{77.69}} & \addtext{83.25} & \addtext{76.74}\\ 
 \hline
 \addtext{ours (word + sent): 1 CA layer} & \addtext{Faster-RCNN (VG)} & \addtext{49.99} & \addtext{76.72} & \addtext{\textbf{83.51}} & \addtext{74.97} \\
  \hline
 \addtext{ours (word + sent): 2 CA layers} & \addtext{Faster-RCNN (VG)} &  \addtext{\textbf{53.80}} & \addtext{76.69} & \addtext{82.28} & \addtext{\textbf{76.78}} \\
 \hline
\end{tabular}
}
\vspace{-5pt}
\end{table}

\paragraph{Discussion.} This work introduces the problem of localizing narrated interactions in uncurated videos without localization supervision. To address this task, we propose a novel Contrastive Multilayer Multimodal Attention module that facilitates repeated attention interactions between spatiotemporal regions and words to learn their latent alignment. Additionally, we introduce a new evaluation dataset that provides spatial annotations for narrated instructions in videos. Evaluations on the three separate tasks of interaction and object localization in videos as well as phrase grounding in images demonstrate the ability of our model to ground both words and phrases more effectively than the shallow and full-attention baselines. Our approach opens up the possibility of further exploration in other modalities. Learning to associate relevant spatial regions with natural language sentences may also be beneficial to learning richer video representations, particularly with the recent gravitation towards self-supervised representation learning.

{\small
\bibliographystyle{ieee_fullname}
\bibliography{egbib}

\begin{thebibliography}{10}\itemsep=-1pt

\bibitem{abnar2020quantifying}
Samira Abnar and Willem Zuidema.
\newblock Quantifying attention flow in transformers.
\newblock {\em arXiv preprint arXiv:2005.00928}, 2020.

\bibitem{afouras2020selfsupervised}
Triantafyllos Afouras, Andrew Owens, Joon~Son Chung, and Andrew Zisserman.
\newblock Self-supervised learning of audio-visual objects from video.
\newblock {\em ECCV}, 2020.

\bibitem{akbari2019multi}
Hassan Akbari, Svebor Karaman, Surabhi Bhargava, Brian Chen, Carl Vondrick, and
  Shih-Fu Chang.
\newblock Multi-level multimodal common semantic space for image-phrase
  grounding.
\newblock In {\em Proceedings of the IEEE/CVF Conference on Computer Vision and
  Pattern Recognition}, pages 12476--12486, 2019.

\bibitem{amrani2020self}
Elad Amrani, Rami Ben-Ari, Inbar Shapira, Tal Hakim, and Alex Bronstein.
\newblock Self-supervised object detection and retrieval using unlabeled
  videos.
\newblock In {\em Proceedings of the IEEE/CVF Conference on Computer Vision and
  Pattern Recognition Workshops}, pages 954--955, 2020.

\bibitem{anne2017localizing}
Lisa Anne~Hendricks, Oliver Wang, Eli Shechtman, Josef Sivic, Trevor Darrell,
  and Bryan Russell.
\newblock Localizing moments in video with natural language.
\newblock In {\em Proceedings of the IEEE international conference on computer
  vision}, pages 5803--5812, 2017.

\bibitem{benaim2020speednet}
Sagie Benaim, Ariel Ephrat, Oran Lang, Inbar Mosseri, William~T Freeman,
  Michael Rubinstein, Michal Irani, and Tali Dekel.
\newblock Speednet: Learning the speediness in videos.
\newblock In {\em Proceedings of the IEEE/CVF Conference on Computer Vision and
  Pattern Recognition}, pages 9922--9931, 2020.

\bibitem{bertasius2020cobe}
Gedas Bertasius and Lorenzo Torresani.
\newblock Cobe: Contextualized object embeddings from narrated instructional
  video.
\newblock {\em arXiv preprint arXiv:2007.07306}, 2020.

\bibitem{opencv_library}
G. Bradski.
\newblock {The OpenCV Library}.
\newblock {\em Dr. Dobb's Journal of Software Tools}, 2000.

\bibitem{chen2018temporally}
Jingyuan Chen, Xinpeng Chen, Lin Ma, Zequn Jie, and Tat-Seng Chua.
\newblock Temporally grounding natural sentence in video.
\newblock In {\em Proceedings of the 2018 conference on empirical methods in
  natural language processing}, pages 162--171, 2018.

\bibitem{chen2020simple}
Ting Chen, Simon Kornblith, Mohammad Norouzi, and Geoffrey Hinton.
\newblock A simple framework for contrastive learning of visual
  representations.
\newblock In {\em International conference on machine learning}, pages
  1597--1607. PMLR, 2020.

\bibitem{chen2020improved}
Xinlei Chen, Haoqi Fan, Ross Girshick, and Kaiming He.
\newblock Improved baselines with momentum contrastive learning.
\newblock {\em arXiv preprint arXiv:2003.04297}, 2020.

\bibitem{chen2019weakly}
Zhenfang Chen, Lin Ma, Wenhan Luo, and Kwan-Yee~K Wong.
\newblock Weakly-supervised spatio-temporally grounding natural sentence in
  video.
\newblock {\em arXiv preprint arXiv:1906.02549}, 2019.

\bibitem{collardvalley}
Collard~Valley Cooks.
\newblock https://www.youtube.com/watch?v=quunrafxqd0.

\bibitem{curiosityculture}
Curiosity Culture.
\newblock https://www.youtube.com/watch?v=jr9qw\textunderscore bjpte.

\bibitem{Damen2020Collection}
Dima Damen, Hazel Doughty, Giovanni~Maria Farinella, Sanja Fidler, Antonino
  Furnari, Evangelos Kazakos, Davide Moltisanti, Jonathan Munro, Toby Perrett,
  Will Price, and Michael Wray.
\newblock The epic-kitchens dataset: Collection, challenges and baselines.
\newblock {\em IEEE Transactions on Pattern Analysis and Machine Intelligence
  (TPAMI)}, 2020.

\bibitem{datta2019align2ground}
Samyak Datta, Karan Sikka, Anirban Roy, Karuna Ahuja, Devi Parikh, and Ajay
  Divakaran.
\newblock Align2ground: Weakly supervised phrase grounding guided by
  image-caption alignment.
\newblock In {\em Proceedings of the IEEE/CVF International Conference on
  Computer Vision}, pages 2601--2610, 2019.

\bibitem{devlin2018bert}
Jacob Devlin, Ming-Wei Chang, Kenton Lee, and Kristina Toutanova.
\newblock Bert: Pre-training of deep bidirectional transformers for language
  understanding.
\newblock {\em arXiv preprint arXiv:1810.04805}, 2018.

\bibitem{fang2015captions}
Hao Fang, Saurabh Gupta, Forrest Iandola, Rupesh~K Srivastava, Li Deng, Piotr
  Doll{\'a}r, Jianfeng Gao, Xiaodong He, Margaret Mitchell, John~C Platt,
  et~al.
\newblock From captions to visual concepts and back.
\newblock In {\em Proceedings of the IEEE conference on computer vision and
  pattern recognition}, pages 1473--1482, 2015.

\bibitem{fernando2017self}
Basura Fernando, Hakan Bilen, Efstratios Gavves, and Stephen Gould.
\newblock Self-supervised video representation learning with odd-one-out
  networks.
\newblock In {\em Proceedings of the IEEE conference on computer vision and
  pattern recognition}, pages 3636--3645, 2017.

\bibitem{girshick2015fast}
Ross Girshick.
\newblock Fast r-cnn.
\newblock In {\em Proceedings of the IEEE international conference on computer
  vision}, pages 1440--1448, 2015.

\bibitem{girshick2014rich}
Ross Girshick, Jeff Donahue, Trevor Darrell, and Jitendra Malik.
\newblock Rich feature hierarchies for accurate object detection and semantic
  segmentation.
\newblock In {\em Proceedings of the IEEE conference on computer vision and
  pattern recognition}, pages 580--587, 2014.

\bibitem{gupta2020contrastive}
Tanmay Gupta, Arash Vahdat, Gal Chechik, Xiaodong Yang, Jan Kautz, and Derek
  Hoiem.
\newblock Contrastive learning for weakly supervised phrase grounding.
\newblock {\em arXiv preprint arXiv:2006.09920}, 2020.

\bibitem{han2020memory}
Tengda Han, Weidi Xie, and Andrew Zisserman.
\newblock Memory-augmented dense predictive coding for video representation
  learning.
\newblock In {\em Computer Vision--ECCV 2020: 16th European Conference,
  Glasgow, UK, August 23--28, 2020, Proceedings, Part III 16}, pages 312--329.
  Springer, 2020.

\bibitem{he2020momentum}
Kaiming He, Haoqi Fan, Yuxin Wu, Saining Xie, and Ross Girshick.
\newblock Momentum contrast for unsupervised visual representation learning.
\newblock In {\em Proceedings of the IEEE/CVF Conference on Computer Vision and
  Pattern Recognition}, pages 9729--9738, 2020.

\bibitem{ioffe2015batch}
Sergey Ioffe and Christian Szegedy.
\newblock Batch normalization: Accelerating deep network training by reducing
  internal covariate shift.
\newblock In {\em International conference on machine learning}, pages
  448--456. PMLR, 2015.

\bibitem{jabri2020space}
Allan Jabri, Andrew Owens, and Alexei~A Efros.
\newblock Space-time correspondence as a contrastive random walk.
\newblock {\em arXiv preprint arXiv:2006.14613}, 2020.

\bibitem{kalogeiton2017action}
Vicky Kalogeiton, Philippe Weinzaepfel, Vittorio Ferrari, and Cordelia Schmid.
\newblock Action tubelet detector for spatio-temporal action localization.
\newblock In {\em Proceedings of the IEEE International Conference on Computer
  Vision}, pages 4405--4413, 2017.

\bibitem{alicia_vid}
Alicia Kirby.
\newblock https://www.youtube.com/watch?v=r-ennr\textunderscore oh8a.

\bibitem{krishna2017visual}
Ranjay Krishna, Yuke Zhu, Oliver Groth, Justin Johnson, Kenji Hata, Joshua
  Kravitz, Stephanie Chen, Yannis Kalantidis, Li-Jia Li, David~A Shamma, et~al.
\newblock Visual genome: Connecting language and vision using crowdsourced
  dense image annotations.
\newblock {\em International journal of computer vision}, 123(1):32--73, 2017.

\bibitem{li2017tracking}
Zhenyang Li, Ran Tao, Efstratios Gavves, Cees~GM Snoek, and Arnold~WM
  Smeulders.
\newblock Tracking by natural language specification.
\newblock In {\em Proceedings of the IEEE Conference on Computer Vision and
  Pattern Recognition}, pages 6495--6503, 2017.

\bibitem{lin2014microsoft}
Tsung-Yi Lin, Michael Maire, Serge Belongie, James Hays, Pietro Perona, Deva
  Ramanan, Piotr Doll{\'a}r, and C~Lawrence Zitnick.
\newblock Microsoft coco: Common objects in context.
\newblock In {\em European conference on computer vision}, pages 740--755.
  Springer, 2014.

\bibitem{loshchilov2017decoupled}
Ilya Loshchilov and Frank Hutter.
\newblock Decoupled weight decay regularization.
\newblock {\em arXiv preprint arXiv:1711.05101}, 2017.

\bibitem{lucas1981iterative}
Bruce~D Lucas, Takeo Kanade, et~al.
\newblock An iterative image registration technique with an application to
  stereo vision.
\newblock Vancouver, British Columbia, 1981.

\bibitem{kharmamedic}
Kharma Medic.
\newblock https://www.youtube.com/watch?v=b\textunderscore zkdjlrghc.

\bibitem{miech2020end}
Antoine Miech, Jean-Baptiste Alayrac, Lucas Smaira, Ivan Laptev, Josef Sivic,
  and Andrew Zisserman.
\newblock End-to-end learning of visual representations from uncurated
  instructional videos.
\newblock In {\em Proceedings of the IEEE/CVF Conference on Computer Vision and
  Pattern Recognition}, pages 9879--9889, 2020.

\bibitem{miech2019howto100m}
Antoine Miech, Dimitri Zhukov, Jean-Baptiste Alayrac, Makarand Tapaswi, Ivan
  Laptev, and Josef Sivic.
\newblock Howto100m: Learning a text-video embedding by watching hundred
  million narrated video clips.
\newblock In {\em Proceedings of the IEEE/CVF International Conference on
  Computer Vision}, pages 2630--2640, 2019.

\bibitem{mikolov2013efficient}
Tomas Mikolov, Kai Chen, Greg Corrado, and Jeffrey Dean.
\newblock Efficient estimation of word representations in vector space.
\newblock {\em arXiv preprint arXiv:1301.3781}, 2013.

\bibitem{oord2019representation}
Aaron van~den Oord, Yazhe Li, and Oriol Vinyals.
\newblock Representation learning with contrastive predictive coding.
\newblock {\em arXiv preprint arXiv:1807.03748}, 2019.

\bibitem{peters2018deep}
Matthew~E Peters, Mark Neumann, Mohit Iyyer, Matt Gardner, Christopher Clark,
  Kenton Lee, and Luke Zettlemoyer.
\newblock Deep contextualized word representations.
\newblock {\em arXiv preprint arXiv:1802.05365}, 2018.

\bibitem{plummer2015flickr30k}
Bryan~A Plummer, Liwei Wang, Chris~M Cervantes, Juan~C Caicedo, Julia
  Hockenmaier, and Svetlana Lazebnik.
\newblock Flickr30k entities: Collecting region-to-phrase correspondences for
  richer image-to-sentence models.
\newblock In {\em Proceedings of the IEEE international conference on computer
  vision}, pages 2641--2649, 2015.

\bibitem{CLIPRadford2021}
Alec Radford, Jong~Wook Kim, Chris Hallacy, Aditya Ramesh, Gabriel Goh,
  Sandhini Agarwal, Girish Sastry, Amanda Askell, Pamela Mishkin, Jack Clark,
  Gretchen Krueger, and Ilya Sutskever.
\newblock Learning transferable visual models from natural language
  supervision, 2021.

\bibitem{yolov3}
Joseph Redmon and Ali Farhadi.
\newblock Yolov3: An incremental improvement.
\newblock {\em arXiv}, 2018.

\bibitem{renNIPS15fasterrcnn}
Shaoqing Ren, Kaiming He, Ross Girshick, and Jian Sun.
\newblock Faster {R-CNN}: Towards real-time object detection with region
  proposal networks.
\newblock In {\em Advances in Neural Information Processing Systems}, 2015.

\bibitem{Sadhu_2020_CVPR}
Arka Sadhu, Kan Chen, and Ram Nevatia.
\newblock Video object grounding using semantic roles in language description.
\newblock In {\em Proceedings of the IEEE/CVF Conference on Computer Vision and
  Pattern Recognition (CVPR)}, June 2020.

\bibitem{selvaraju2020casting}
Ramprasaath~R Selvaraju, Karan Desai, Justin Johnson, and Nikhil Naik.
\newblock Casting your model: Learning to localize improves self-supervised
  representations.
\newblock {\em arXiv preprint arXiv:2012.04630}, 2020.

\bibitem{Shan20}
Dandan Shan, Jiaqi Geng, Michelle Shu, and David Fouhey.
\newblock Understanding human hands in contact at internet scale.
\newblock 2020.

\bibitem{shi2019not}
Jing Shi, Jia Xu, Boqing Gong, and Chenliang Xu.
\newblock Not all frames are equal: Weakly-supervised video grounding with
  contextual similarity and visual clustering losses.
\newblock In {\em Proceedings of the IEEE Conference on Computer Vision and
  Pattern Recognition}, pages 10444--10452, 2019.

\bibitem{sixsisters}
Six~Sisters' Stuff.
\newblock https://www.youtube.com/watch?v=vic2l-iyrq0.

\bibitem{sun2019cbt}
Chen Sun, Fabien Baradel, Kevin Murphy, and Cordelia Schmid.
\newblock Contrastive bidirectional transformer for temporal representation
  learning.
\newblock In {\em arXiv}, June 2019.

\bibitem{sun2019learning}
Chen Sun, Fabien Baradel, Kevin Murphy, and Cordelia Schmid.
\newblock Learning video representations using contrastive bidirectional
  transformer.
\newblock {\em arXiv preprint arXiv:1906.05743}, 2019.

\bibitem{uijlings2013selective}
Jasper~RR Uijlings, Koen~EA Van De~Sande, Theo Gevers, and Arnold~WM Smeulders.
\newblock Selective search for object recognition.
\newblock {\em International journal of computer vision}, 104(2):154--171,
  2013.

\bibitem{vaswani2017attention}
Ashish Vaswani, Noam Shazeer, Niki Parmar, Jakob Uszkoreit, Llion Jones,
  Aidan~N Gomez, Lukasz Kaiser, and Illia Polosukhin.
\newblock Attention is all you need.
\newblock {\em arXiv preprint arXiv:1706.03762}, 2017.

\bibitem{wang2020selfsupvideo}
Jiangliu Wang, Jianbo Jiao, and Yun-Hui Liu.
\newblock Self-supervised video representation learning by pace prediction.
\newblock In Andrea Vedaldi, Horst Bischof, Thomas Brox, and Jan-Michael Frahm,
  editors, {\em Computer Vision -- ECCV 2020}, 2020.

\bibitem{wang2019without}
Josiah Wang and Lucia Specia.
\newblock Phrase localization without paired training examples.
\newblock In {\em Proceedings of the IEEE international conference on computer
  vision}, 2019.

\bibitem{wei2018learning}
Donglai Wei, Joseph~J Lim, Andrew Zisserman, and William~T Freeman.
\newblock Learning and using the arrow of time.
\newblock In {\em Proceedings of the IEEE Conference on Computer Vision and
  Pattern Recognition}, pages 8052--8060, 2018.

\bibitem{smokingrillin}
Smokin' \&~Grillin' wit AB.
\newblock https://www.youtube.com/watch?v=\textunderscore wvcrj15s-m.

\bibitem{woolworths_vid}
Woolworths.
\newblock https://www.youtube.com/watch?v=viwpmylgps0.

\bibitem{xie2018s3d}
Saining Xie, Chen Sun, Jonathan Huang, Zhuowen Tu, and Kevin Murphy.
\newblock Rethinking spatiotemporal feature learning: Speed-accuracy trade-offs
  in video classification.
\newblock In {\em European Conference on Computer Vision}, 2018.

\bibitem{xu2017r}
Huijuan Xu, Abir Das, and Kate Saenko.
\newblock R-c3d: Region convolutional 3d network for temporal activity
  detection.
\newblock In {\em Proceedings of the IEEE international conference on computer
  vision}, pages 5783--5792, 2017.

\bibitem{yang2019xlnet}
Zhilin Yang, Zihang Dai, Yiming Yang, Jaime Carbonell, Ruslan Salakhutdinov,
  and Quoc~V Le.
\newblock Xlnet: Generalized autoregressive pretraining for language
  understanding.
\newblock {\em arXiv preprint arXiv:1906.08237}, 2019.

\bibitem{zhang2020does}
Zhu Zhang, Zhou Zhao, Yang Zhao, Qi Wang, Huasheng Liu, and Lianli Gao.
\newblock Where does it exist: Spatio-temporal video grounding for multi-form
  sentences.
\newblock In {\em Proceedings of the IEEE/CVF Conference on Computer Vision and
  Pattern Recognition}, pages 10668--10677, 2020.

\bibitem{zhao2020distilling}
Nanxuan Zhao, Zhirong Wu, Rynson~WH Lau, and Stephen Lin.
\newblock Distilling localization for self-supervised representation learning.
\newblock {\em arXiv preprint arXiv:2004.06638}, 2020.

\bibitem{zhou2019grounded}
Luowei Zhou, Yannis Kalantidis, Xinlei Chen, Jason~J Corso, and Marcus
  Rohrbach.
\newblock Grounded video description.
\newblock In {\em CVPR}, 2019.

\bibitem{ZhLoCoBMVC18}
Luowei Zhou, Nathan Louis, and Jason~J Corso.
\newblock Weakly-supervised video object grounding from text by loss weighting
  and object interaction.
\newblock In {\em British Machine Vision Conference}, 2018.

\bibitem{youcook2zhou2018}
Luowei Zhou, Chenliang Xu, and Jason Corso.
\newblock Towards automatic learning of procedures from web instructional
  videos.
\newblock In {\em Proceedings of the AAAI Conference on Artificial
  Intelligence}, volume~32, 2018.

\end{thebibliography}
}

\clearpage
\appendix
\section{Supplementary}
In this supplementary material, we provide the following additions to the main submission:
\begin{enumerate}
    \item[A.1.] weakly supervised phrase grounding experiments
        \begin{enumerate}
            \item implementation details
            \item hybrid model
            \item ablation results
        \end{enumerate}
    \item[A.2.] grounding narrated instructions in videos experiments
        \begin{enumerate}
            \item self-training dataset
            \item implementation details of baselines
            \item mathematical formalization of cross-attention layer
            \item additional details about our model architecture
            \item Active Hands model implementation and results
            \item limitations
        \end{enumerate}
    \item[A.3.] details about our YouCook2-Interactions dataset
        \begin{enumerate}
            \item description of the annotation process
            \item the guiding principles for frame relevance labeling 
            \item the guiding principles for bounding box annotation 
        \end{enumerate}
\end{enumerate}

\subsection{Weakly supervised phrase grounding experiments} \label{sec: appendix-phrase-grounding}
\paragraph{Implementation details.}  Following Gupta \emph{et al.} \cite{gupta2020contrastive}, we extract image region features with a Faster-RCNN~\cite{renNIPS15fasterrcnn} model pretrained on Visual Genome \cite{krishna2017visual} and contextualized word embeddings from a pretrained BERT \cite{devlin2018bert} model. The text value projection is implemented as a two-layer MLP with Batch Normalization \cite{ioffe2015batch}. We use ReLU as the activation function. We set the joint embedding space dimension to the same as Gupta \emph{et al}.~\cite{gupta2020contrastive} (d=384) and use a batch size of 50 images. We adopt a constant learning rate of 1e-5 and the Adam optimizer. Our approach is built on the publicly available implementation of Gupta \emph{et al}.~\cite{gupta2020contrastive}. Finally, we train our model for 8 hours on a single V100 GPU.

\begin{figure}
    \centering
    \subfloat[\centering \textbf{Gupta \emph{et al.}} \cite{gupta2020contrastive}] {{\includegraphics[width=0.45\columnwidth]{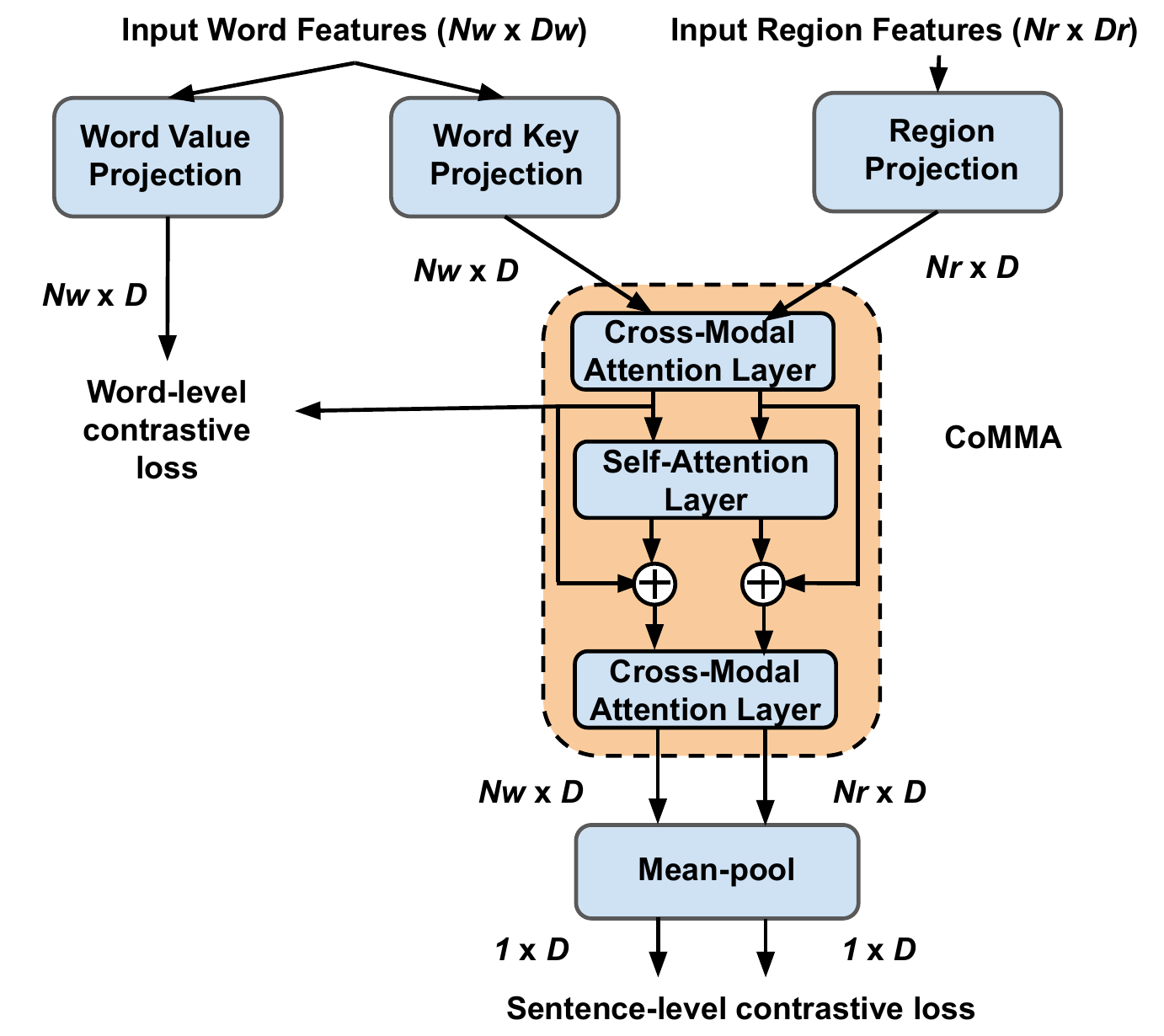} } \label{fig:image-orig-model}}
    \qquad
    \subfloat[\centering \textbf{Hybrid model for weakly supervised phrase grounding.}]
    {{\includegraphics[width=0.45\columnwidth]{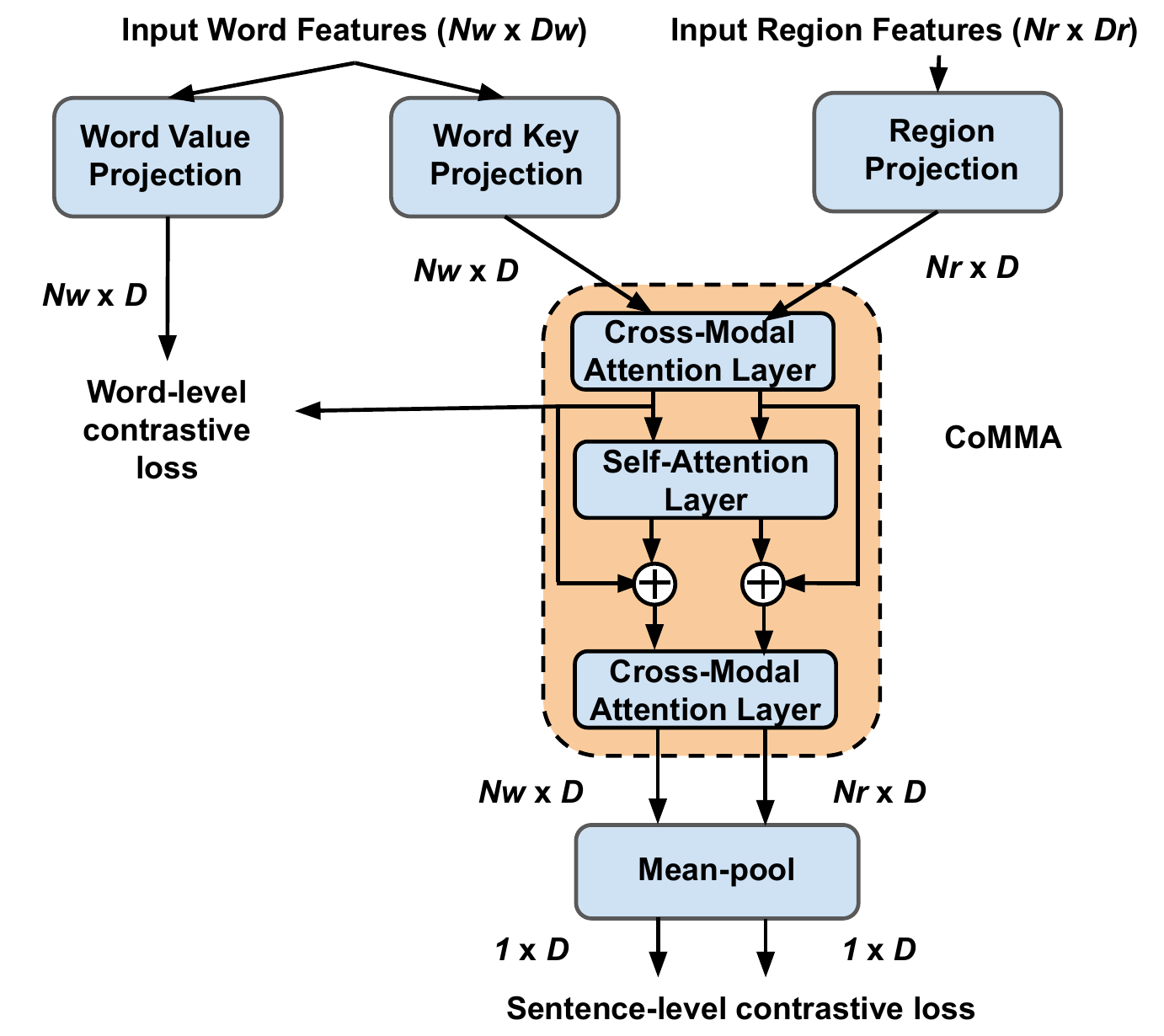} } \label{fig:image-hybrid-model}}
    \caption{$N_w$, $N_r$, $D_w$, $D_r$ and $D$ denote the number of words, regions as well as the dimensions of the input word and region features and the joint embedding space, respectively.}
    \label{fig:phrase-grounding-models}
\end{figure}

\paragraph{Hybrid model.} We provide an illustration of our weakly supervised phrase grounding model in Figure~\ref{fig:image-hybrid-model} (this supplemental). $D_w$ and $D_r$ denote the dimensions of the input word and region features while $N_w$ and $N_r$ denote the number of words and regions, respectively. We use $D$ to indicate the dimension of the joint embedding space. To incorporate our proposed CoMMA into the model of Gupta \emph{et al}.~\cite{gupta2020contrastive}  (Figure~\ref{fig:image-orig-model} (this supplemental)), we begin by removing a value projection MLP for the region features. This is due to the fact that our proposed CoMMA module does not project the region features to different subspaces for computing cross-attention. The model of Gupta \emph{et al}.~\cite{gupta2020contrastive} is illustrated in Figure~\ref{fig:image-orig-model} (this supplemental). The outputs of the region and word key projection MLPs are passed as inputs into our cross-attention layer which computes contextualized region and word representations, respectively. We use the final contextualized word representations to compute the sentence-level loss. We found that using the contextualized word representation outputs from the first cross-attention layer to compute the word-level contrastive loss is critical for optimizing the model successfully. Finally, the sentence loss is weighted by a hyperparameter. We use the attention mask from the first cross-attention layer during evaluation.

\paragraph{Negative Noun Loss.} In our experiments, we also adopt a negative noun loss as used in Gupta \emph{et al}.~\cite{gupta2020contrastive}. Specifically, we create context-preserving negative captions for an image by substituting a noun in its original caption with negative nouns, that are sampled from a pretrained BERT \cite{devlin2018bert} model. We compute the negative noun loss by contrasting the image region representations against these negative captions. For the rest of this supplementary, we use the term `word-level contrastive loss' to include the negative noun loss as well.

\paragraph{Ablation Results.} Table ~\ref{image-sentence-loss} (this supplemental) reports ablation results of the layers in our proposed CoMMA module when trained only with the sentence-level contrastive loss. In this case, adding cross-attention and self-attention layers generally helps to improve localization performance. As mentioned above, we remove the value projection for region features from the model of Gupta \emph{et al}.~\cite{gupta2020contrastive} to incorporate the CoMMA module. This modification results in an initial drop in performance when a single cross-modal attention layer is used, as shown in the first row of Table ~\ref{table:sentence-loss-addition} (this supplemental) and Table ~\ref{table:image-phrase-grounding} (main paper). We also observe that the sentence-level contrastive loss is complementary to the word-level contrastive loss (on its own, it performs reasonably well). For example, in the case where only one cross-attention layer is used, adding the sentence-level contrastive loss leads to a 2.5\% in the R@1 accuracy. However, using the complete CoMMA model hurts performance. We hypothesize that it is overfitting on the smaller dataset and we have not found the optimal optimization setting. Finally, Table ~\ref{table:sentence-loss-weight-ablation} (this supplemental) reports the ablation results over the different weight values of the sentence-level contrastive loss. We achieve the best performance when we use a weight value of $\lambda = 0.005$ with two cross-attention layers.

\begin{table}
\centering
 {
\caption{
{\bf CoMMA component ablation on Flickr30k with only sentence-level loss.} We observe that adding cross-attention and self-attention layers generally help to improve the localization accuracy across the Recall@K criteria.
}
 \begin{tabular}{| c | c | c | c| c | c| c|} 
 \hline
Cross-Modal  & Cross-Modal  & Self- & R@1 & R@5 & R@10 & Pt Recall\\
Attention 1. & Attention 2 & Attention & & & & \\
 \hline
\ding{51} & \ding{55} & \ding{55} & \addtext{35.43} & \addtext{64.48} & \addtext{73.92} & \addtext{68.68}\\ 
 \hline
 \ding{51} & \ding{51} & \ding{55} & \addtext{39.03} & \addtext{66.91} & \addtext{75.15} & \addtext{\textbf{70.13}} \\
  \hline
  \ding{51} &  \ding{51} &  \ding{51} & \addtext{\textbf{39.31}} & \addtext{\textbf{68.43}} & \addtext{\textbf{77.35}} & \addtext{67.36} \\
 \hline
\end{tabular} \label{image-sentence-loss}
}
\end{table}

\begin{table}
\centering
\caption{
{\bf Adding sentence loss to our hybrid model on Flickr30k.} We observe that the sentence loss is complementary to the word loss (on its own, it performs reasonably well). We also note that adding more attention layers also helps to improve localization performance.
} \label{table:sentence-loss-addition}
 {
 \begin{tabular}{| c | c | c | c| c | c| c| c|} 
 \hline
Cross-Model  & Cross-Modal & Self- & Sentence  & R@1 & R@5 & R@10 & Pt \\
Attention 1 & Attention 2 & Attention & Loss & & & & Recall \\
 \hline
\ding{51} & \ding{55} & \ding{55} & \addtext{No} & \addtext{49.99} & \addtext{\textbf{76.72}} & \addtext{\textbf{83.51}} & \addtext{74.97}\\ 
 \hline
 \ding{51} & \ding{55} & \ding{55} & \addtext{Yes} & \addtext{52.43} & \addtext{76.11} & \addtext{81.75} & \addtext{76.65}\\ 
 \hline
 \ding{51} & \ding{51} & \ding{55} & \addtext{Yes} & \addtext{\textbf{53.80}} & \addtext{76.69} & \addtext{82.28} & \addtext{\textbf{76.78}} \\
  \hline
  \ding{51} &  \ding{51} &  \ding{51} & \addtext{Yes} & \addtext{48.53} & \addtext{76.26} & \addtext{82.98} & \addtext{73.78} \\
 \hline
\end{tabular}
}
\end{table}

\begin{table}
\centering
\caption{
{\bf Sentence loss weight ablation with our hybrid model on Flickr30k.} The best performance is obtained with two cross-attention layers with the weight of the sentence loss set to 0.005. In the case where only one cross-attention layer is used, increasing the value of $\lambda$ from 0 to 0.01 or 0.005 leads to a 2\% in the R@1 accuracy.
}
 {
 \begin{tabular}{| c | c | c | c| c | c| c| c|} 
 \hline
Cross-Model  & Cross-Modal & Self- & Sentence  & R@1 & R@5 & R@10 & Pt \\
Attention 1 & Attention 2 & Attention & Loss Weight $\lambda$ & & & & Recall \\
 \hline
\ding{51} & \ding{55} & \ding{55} & \addtext{0.0} & \addtext{49.99} & \addtext{\textbf{76.72}} & \addtext{\textbf{83.51}} & \addtext{74.97}\\ 
 \hline
\ding{51} & \ding{55} & \ding{55} & \addtext{0.01} & \addtext{52.59} & \addtext{75.42} & \addtext{80.82} & \addtext{75.80}\\ 
 \hline
 \ding{51} & \ding{55} & \ding{55} & \addtext{0.005} & \addtext{52.43} & \addtext{76.11} & \addtext{81.75} & \addtext{76.65}\\ 
 \hline
 \ding{51} & \ding{55} & \ding{55} & \addtext{0.001} & \addtext{45.72} & \addtext{71.05} & \addtext{77.84} & \addtext{71.99}\\ 
 \hline
 \ding{51} & \ding{51} & \ding{55} & \addtext{0.01} & \addtext{52.44} & \addtext{75.55} & \addtext{81.13} & \addtext{75.71} \\
  \hline
  \ding{51} & \ding{51} & \ding{55} & \addtext{0.005} & \addtext{\textbf{53.80}} & \addtext{76.69} & \addtext{82.28} & \addtext{\textbf{76.78}} \\
  \hline
  \ding{51} & \ding{51} & \ding{55} & \addtext{0.01} & \addtext{46.23} & \addtext{73.32} & \addtext{80.70} & \addtext{72.94} \\
  \hline
  \ding{51} &  \ding{51} &  \ding{51} & \addtext{0.01} & \addtext{47.57} & \addtext{75.59} & \addtext{82.63} & \addtext{72.92} \\
   \hline
  \ding{51} &  \ding{51} &  \ding{51} & \addtext{0.005} & \addtext{48.53} & \addtext{76.26} & \addtext{82.98} & \addtext{73.78} \\
   \hline
  \ding{51} &  \ding{51} &  \ding{51} & \addtext{0.01} & \addtext{48.81} & \addtext{76.33} & \addtext{82.99} & \addtext{75.32} \\
 \hline
\end{tabular}
}
\label{table:sentence-loss-weight-ablation}
\end{table}

\subsection{Grounding narrated instructions in videos}

\paragraph{Self-training dataset} The HowTo100M \cite{miech2020end} dataset contains approximately 1.2 million instructional videos that are sourced from YouTube, encompassing multiple domains such as cooking and hand-crafting. These videos contain transcribed narrations that are either uploaded manually by users or are the output of an automatic speech recognition (ASR) system. Furthermore, the narrations have been pre-processed to remove all stop words. As noted in prior work~\cite{miech2020end}, there is a weak temporal alignment between the narrations and the video clips, adding another degree of difficulty in training with such videos. The misalignments are due to the nature of instructional videos where an interaction is often narrated before or after demonstrating it.

\paragraph{Implementation details.} To avoid the time-consuming and computationally expensive process of pretraining the feature encoders used as inputs to our model, we leverage the publicly available pretrained weights of a S3DG model \cite{miech2020end} trained with self-supervision. We use Word2vec embeddings \cite{mikolov2013efficient} to represent semantic information contained by words in the narrations. Note that our proposed model is still fully self-supervised since the pretraining step for both the video and language encoders is also performed on the HowTo100M dataset without curated temporal or language annotations. The language encoder leverages pretrained Word2Vec~\cite{mikolov2013efficient} embeddings. In our experiments, we set the dimension of the joint embedding space $D$ to be 512. We use identity and linear projections with the same input and output dimensions to represent $W_K$, $W_Q$ and $W_V$ in cross-attention and self-attention layers, respectively. To train our proposed model, we set a learning rate of 1e-4 and optimize the model using the AdamW \cite{loshchilov2017decoupled} optimizer. At the beginning of training, we apply linear warmup for one epoch. We train our model on 8 V100 GPUs which takes about a day. During inference, we adopt the practice in prior work where the resolution of the input frame is set to 224 x 224. Each input video clip consists of 16 frames. The dimensions of the final feature map are T x H x W x D = 2 x 4 x 4 x 512, where T, H, W, D correspond to the temporal, height, width, and feature dimensions.

\paragraph{Mathematical formalization of the cross-attention layer}
We provide a more detailed formulation of the cross-attention layer in our CoMMA module here. In particular, we explain how we prevent the mixing of features from different modalities in this layer. Let us first consider an example where we have two features from different modalities $x_1$ and $x_2$. Let $x$ be the concatenation of the two modalities' features $x = [x_1;x_2]$. We denote the operation in a linear layer as $Ax$ for a matrix $A$. Let us write $A = [A_1 \ A_2]$. Then, $Ax = A_1 x_1 + A_2 x_2$. In this case, the two modalities' features are fused since there is a summation operation between them. 

Now, let us consider our setup where we have sets of "values" features $V_1$ and $V_2$ for two modalities. When full attention is used as in Figure ~\ref{model_fig}c (bottom), the output of the softmax function in Equation (\ref{attn_func}) is a full weight matrix $A = [A_{11} \ A_{12}; A_{21} \ A_{22}]$. When multiplying with the value matrix $V = [V_1 \ V_2]$ in Equation (\ref{attn_func}), the output is $[V_1 A_{11} + V_2 A_{21} \ V_1 A_{21} + V_2 A_{22} ]$. Note that in this case, the feature sets for the two modalities ($V_1$ and $V_2$) are summed together.

When cross-attention is used as in Figure ~\ref{model_fig}c (top-left), the output of the softmax function in Equation (\ref{attn_func}) is the matrix $A = [0 \ A_{12}; A_{21} \ 0]$. When $A$ is multiplied by the value matrix $V$, the output is $[V_2 A_{21} \ V_1 A_{12}]$. Consequently, the features for the two modalities are not summed together. The attention interaction between the two modalities only occurs when computing the softmax weights $A_{12}$ and $A_{21}$. We emphasize that these softmax weights are not features themselves, but instead are used for softly selecting features in the two modalities.

\paragraph{Implementation details of baseline architectures.}
Section \ref{cross_modal_baselines} (main paper) describes existing state-of-the-art cross-modal attention modules. We provide additional implementation details of the full-attention baseline that uses a sentinel vector \cite{sun2019cbt} and the shallow attention module \cite{akbari2019multi}. The full-attention model differs from our proposed CoMMA module in two major ways. First, it computes attention weights over the entire multimodal sequence of words and clip features (full attention) instead of between modalities. Second, it uses a sentinel vector to aggregate information between visual and language features. The sentinel vector is a learnable parameter that has a similar purpose to the classification token that is commonly used in BERT language models \cite{devlin2018bert}. Similar to its counterpart in the BERT model, the sentinel vector is prepended to the concatenated sequence of words and visual features and passed through a transformer to aggregate visual-semantic context in it.\newline

The shallow attention module in Akbari \emph{et al}. \cite{akbari2019multi} computes a shallow cross-modal attention between words and convolutional features from different levels of an image encoder. The shallowness of its model stems from the fact that it only consists of a single layer of cross-modal attention between the language and visual features. In contrast, CoMMA comprises of alternating layers of cross-modal as well as self-attention layers. Note that this is different from the CBT model since it does not use a full attention module.  As evidenced by our results, using a deeper cross-modal attention module improves the ability of our model to find the latent alignment between the language and visual modalities.

Finally, in Table ~\ref{table:object-localization} of the main paper, we note that we were unable to exactly reproduce the reported NAFAE results as the original base features were not available. Our best attempt with newer features was 4\% lower than the reported bounding box detection accuracy. The final pointing accuracy obtained with the new features is 41.65\%. Thus, we report the official higher bounding box detection accuracy of 46.95\%. Our approach significantly outperforms this baseline.

\paragraph{Active Hands experiment implementations and results} We compare our approach to a strongly supervised baseline that localizes \emph{active hands} and interacted objects \cite{Shan20} in Table ~\ref{table:active-hands}. We evaluate their pretrained model on YouCook2-Interactions under 3 settings: 
\begin{enumerate}
    \item taking the center pixel of the detected interacted object bounding box
    \item taking the midpoint of the detected left and right hand bounding boxes
    \item using a combination of the (1) and (2). In the case where no interacted objects are detected, we use (2) alone.
\end{enumerate}

The high performance obtained by the pretrained Active Hands model is expected since hands are very often used in interactions, especially in the domain of instructional videos. However, they also suggest that simply detecting hands will likely not solve the proposed task. In particular, it will be challenging to resolve the ambiguity in scenes with multiple interactions involving different people that are occurring simultaneously. Additionally, this approach is not able to work well if there are erroneous detections of hands or objects or multiple pairs of hands visible. This is especially relevant in footage where the hands are small, making recognition based on local appearance difficult. Furthermore, detecting interacted objects using detected hands alone may not be feasible if the described interaction occurs between two objects without any human actors.

One of the main challenges with using the Active Hands model is that there are different heuristics that can be used to evaluate the localization accuracy, as reflected by the 3 settings described above. More importantly, this approach relies on large-scale bounding box annotations of hands during training and evaluation. In contrast, our proposed approach learns to localize such interactions from publicly available data without relying on such hand-annotated bounding boxes. Also, we would like to emphasize the generalizability of our proposed approach to the phrase grounding task in images that does not involve prominent human actors. We hypothesize that using an active hand detector will not generalize as well as our approach to this task given its reliance on finding noun phrases (instead of interactions) in images. Finally, we note that, in theory, it is possible to apply the CoMMA module on top of the hand and object bounding boxes to potentially improve the performance on the task in \cite{Shan20}.

\begin{table}
\centering
 {
\caption{
{\bf Interaction localization evaluation on YouCook2-Interactions.} We compare our approach against the pretrained strongly supervised Active Hands \cite{Shan20} model under 3 different settings.
}
\label{table:active-hands}
 \begin{tabular}{| c | c |} 
 \hline
 Approach & Localization Accuracy (\%)\\
 \hline
 Setting 1 & 70.40 \\
 \hline
 Setting 2 & 61.47 \\
 \hline
 Setting 3 & 64.47 \\
 \hline
 CoMMA (Ours) & \textbf{55.80}\\
 \hline
\end{tabular}
}
\vspace{-10pt}
\end{table}

\paragraph{Limitations.} \addtext{One limitation is our method operates over coarse feature maps. Future work can aim to refine the localization predictions to compute precise masks for the interactions.
}

\subsection{YouCook2-Interactions}
Due to a lack of suitable datasets for localizing narrated interactions, we introduce an evaluation dataset, YouCook2-Interactions, that provides spatial bounding box annotations for interactions that are described by natural language sentences. Our dataset is built on the validation split of the YouCook2 dataset~\cite{youcook2zhou2018}. The original YouCook2 dataset contains around 2000 videos sourced from YouTube that cover approximately 89 recipes. It provides temporal annotations for segments and accompanying natural language descriptions. Each temporal segment is denoted by its start and end times and is described by a natural language description. Each description contains at least one interaction involving multiple objects (e.g., cut the chicken into cubes and add them to a pan).

\subsubsection{Annotations} \label{data_annotation}
Our annotation process is split into three stages, which we will describe below. In the first stage (Person filtering), we filter frames using an off-the-shelf person detector.  Next (Frame relevant labeling), we label each filtered frame as `relevant' or `irrelevant'. Finally (Bounding box annotations), we annotate each relevant frame with a spatial bounding box. We use qualified workers from the Amazon Mechanical Turk platform for both frame labeling and bounding box annotations.

\paragraph{Person filtering.} 
The primary focus of our dataset is on localizing narrated interactions. As such, frames which do not contain any instances of a person are removed to reduce the number of frames for labeling and annotation. To begin with, we extract all of the frames in the relevant segments from the validation split and run the YoloV3 object detector~\cite{yolov3} on all of the frames to detect instances of humans. If a person is not detected in a given frame, it is removed from the labeling and annotation stages. Close-up frames that are visually relevant to narrated actions are not filtered out as long the object detector detects part of a person such as a hand.

\paragraph{Frame relevance labeling.}
One crucial characteristic of these narrations is that each specified interaction is only relevant to certain parts of the temporal segment. Unlike existing datasets which simply aim to track a referred object across frames, the specified interactions may not be visually relevant throughout the entire segment.  Consequently, in the first stage, we ask workers to label if frames in a provided segment are relevant to any interactions specified within the description. We define people to be interacting with objects if they use their hands to perform an action involving the objects such as stirring or cutting. For each frame, three workers annotate whether the it is relevant and we select the majority label as the ground-truth label.

\paragraph{Bounding box annotations.}
After the set of relevant frames from each segment has been determined from the first stage of annotation, workers are tasked to draw a bounding box around the relevant region on a given frame that corresponds to the described interactions. We provide a general overview of the guiding principles here. See below for the entire set of principles. Since the primary objective of our approach is to localize interactions, the ideal bounding box annotation would enclose the region where the objects are being interacted with according to the natural language description. In cases where multiple objects are mentioned in the narration, in a given frame, only those that are being interacted with should be enclosed. Workers are also asked to enclose  the entire hand and the object if the object is partially occluded in a frame.

\noindent The following sections include the guiding principles given to the annotators that are used in both the frame relevance labeling and bounding box annotation stages. For both stages, annotations are done on frames that are extracted at 1 frame per second (fps) and workers are asked to watch the entire clip to understand the context before they start annotating.

\subsubsection{Frame Relevance Labeling Guidelines}
In the original temporal segment annotations provided in the YouCook2 dataset, we observed that the proportion of relevant frames within the specified temporal segments that corresponds to the described interactions is relatively small. To ensure that only relevant frames have bounding box annotations, the first step in our annotation process tasks workers to watch a given video clip and label if the frames are relevant. In this task, we define people to be interacting with objects if they use their hands to perform an action involving the objects such as stirring or cutting. Please note that if a person is just looking or pointing at the object, it is \emph{NOT} considered as an ‘interaction’.

A frame is labeled as `\textbf{irrelevant}' if it fulfills one of the following conditions:
\begin{enumerate}
    \item it depicts a scene where the person would describe the objects or ingredients being used but the specified actions are not being carried out.
    \item some sentences and their interactions may be ambiguous (\emph{e.g}., boiling water, bake the pizza in the oven). Frames where the person does not interact with the specified objects should be labeled as ‘irrelevant’.
    \item if the action is observed but the objects being interacted with are not visible at all. However, if parts of the objects are visible, then frame should be labeled as ‘relevant’ (Figure \ref{fig:bb_guideline_1}) (this supplemental).
\end{enumerate}

\begin{figure}
    \centering
    \includegraphics[width=\columnwidth]{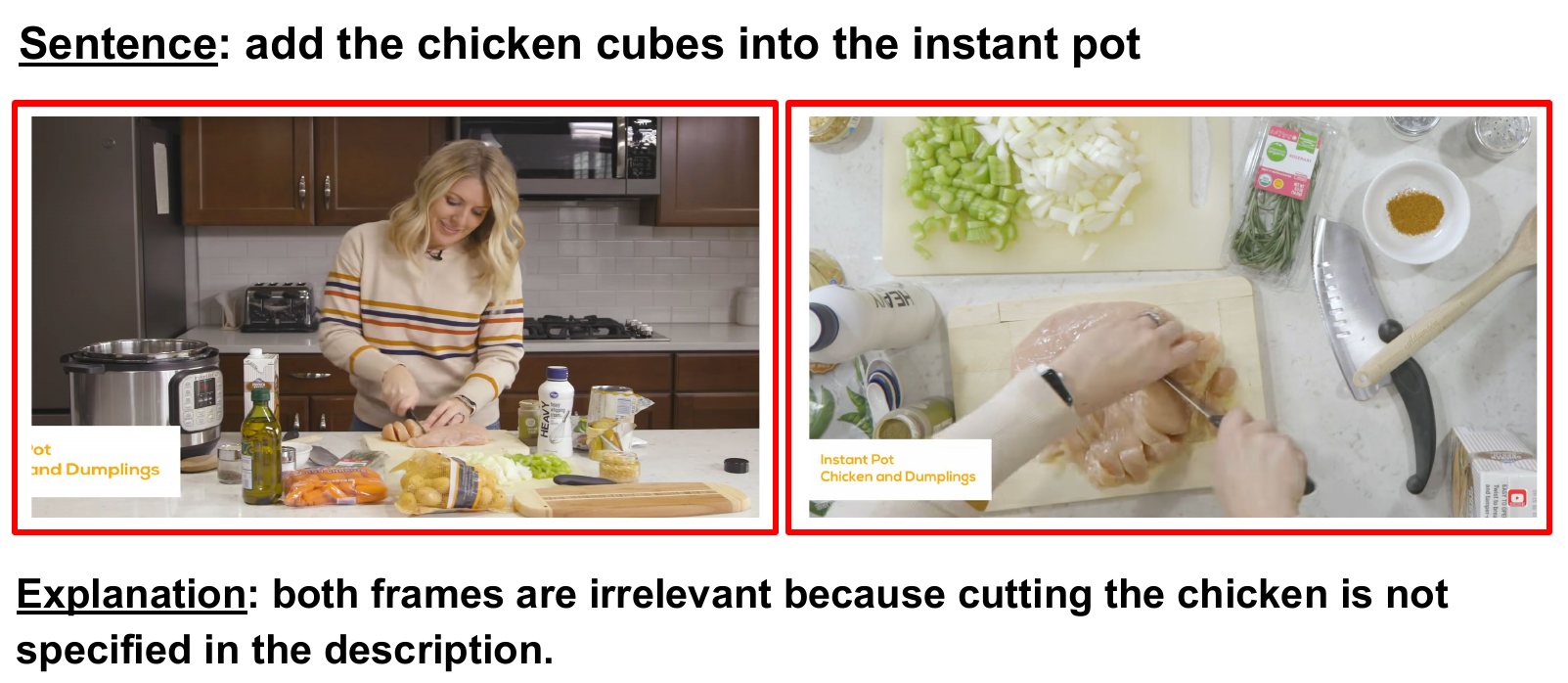}
    \caption{
    Visual example of labeling frames as irrelevant because the sentence is not visually relevant. (Video credit: Six Sisters' Stuff~\cite{sixsisters})
    }
    \label{fig:label_example_1}
\end{figure}

\begin{figure}
    \centering
    \includegraphics[width=\columnwidth]{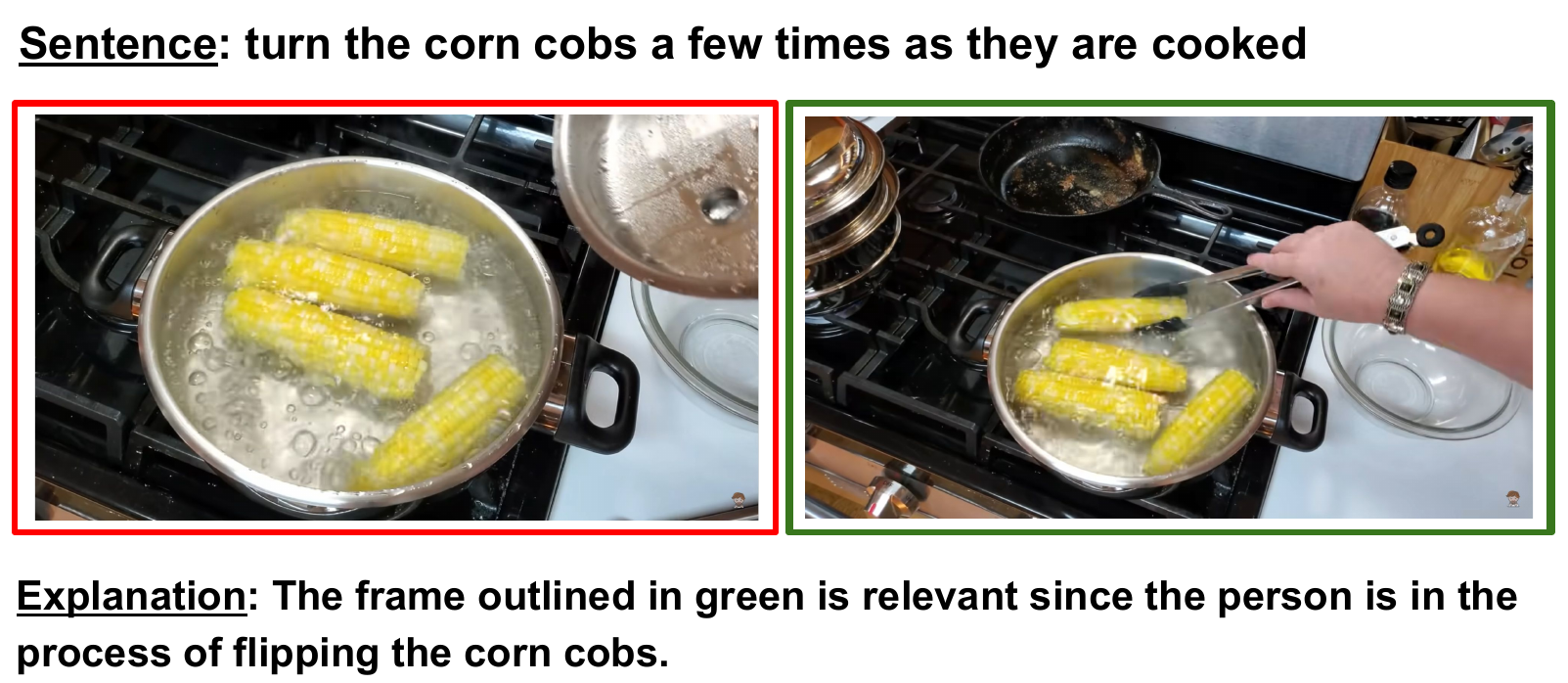}
    \caption{
    Visual example of labeling frames based on their visual relevance to the sentence. (Video credit: Collard Valley Cooks~\cite{collardvalley})
    }
    \label{fig:label_example_2}
\end{figure}

\begin{figure}
    \centering
    \includegraphics[width=\columnwidth]{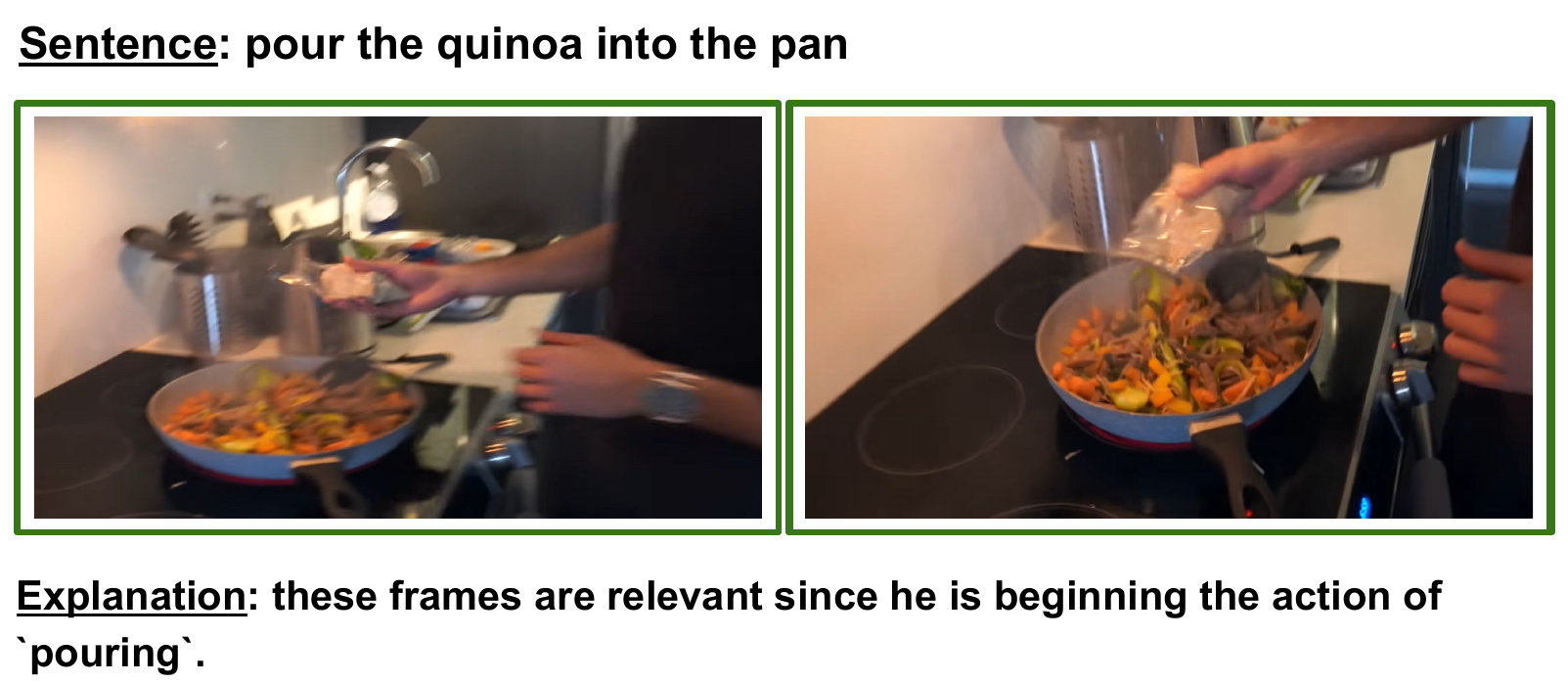}
    \caption{
    Visual example of labeling frames based on their visual relevance to the sentence. (Video credit: Kharma Medic~\cite{kharmamedic})
    }
    \label{fig:label_example_3}
\end{figure}

As noted, there are ambiguous actions like ‘cooking chicken in a pan’, ‘boiling water’ or ‘baking pizza in an oven’. In such cases, all frames with interacted objects as the action is being carried out should be labeled ‘relevant’. Frames where the person does not interact with the object should be labeled as ‘irrelevant’. We provide several visual examples with reasoning for their labels in Figures \ref{fig:label_example_1}, \ref{fig:label_example_2} and \ref{fig:label_example_3} (this supplemental). The frames that are outlined in red and green are marked as `irrelevant` and `relevant`, respectively.

\subsubsection{Bounding Box Annotation Guidelines}
\begin{figure}
    \centering
    \includegraphics[width=\columnwidth]{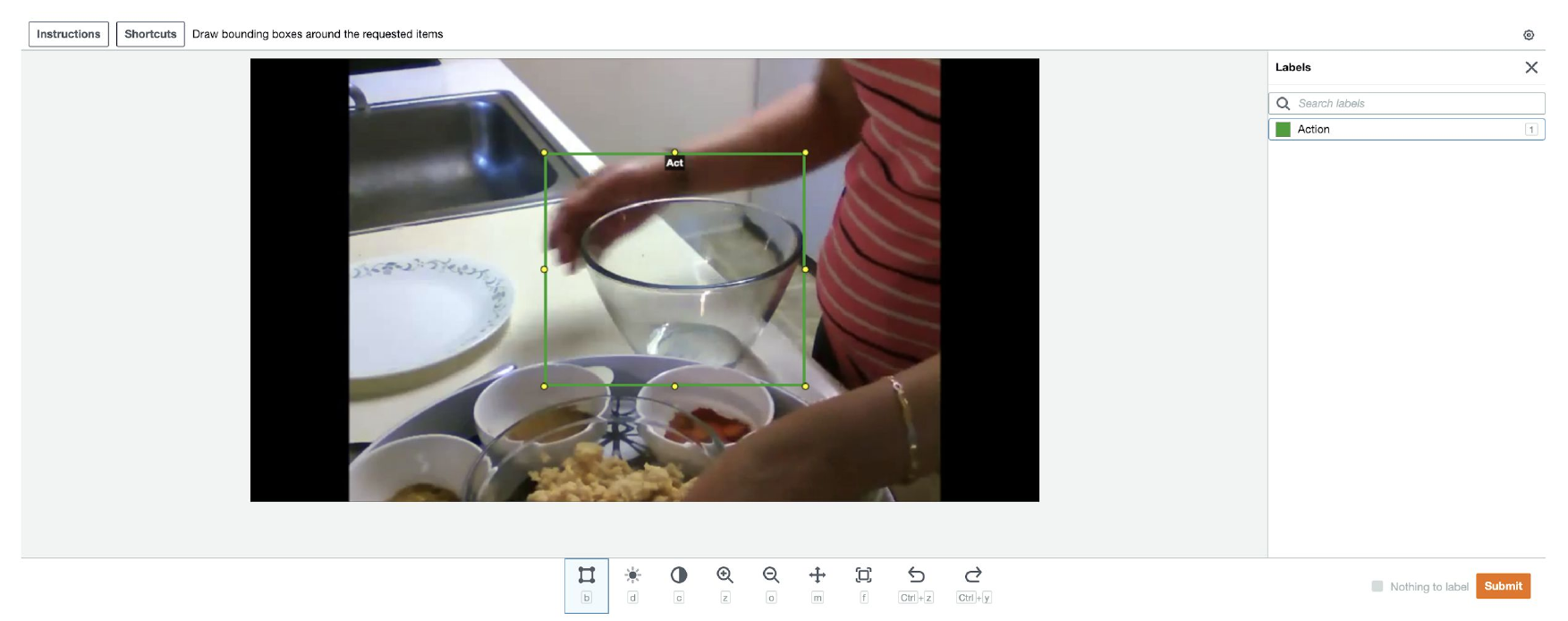}
    \caption{
    An example of the bounding box annotation interface used on Amazon Mechanical Turk. (Video credit: Curiosity Culture~\cite{curiosityculture})
    }
    \label{fig:annotation_interface}
\end{figure}

\begin{figure}
    \centering
    \includegraphics[width=\columnwidth]{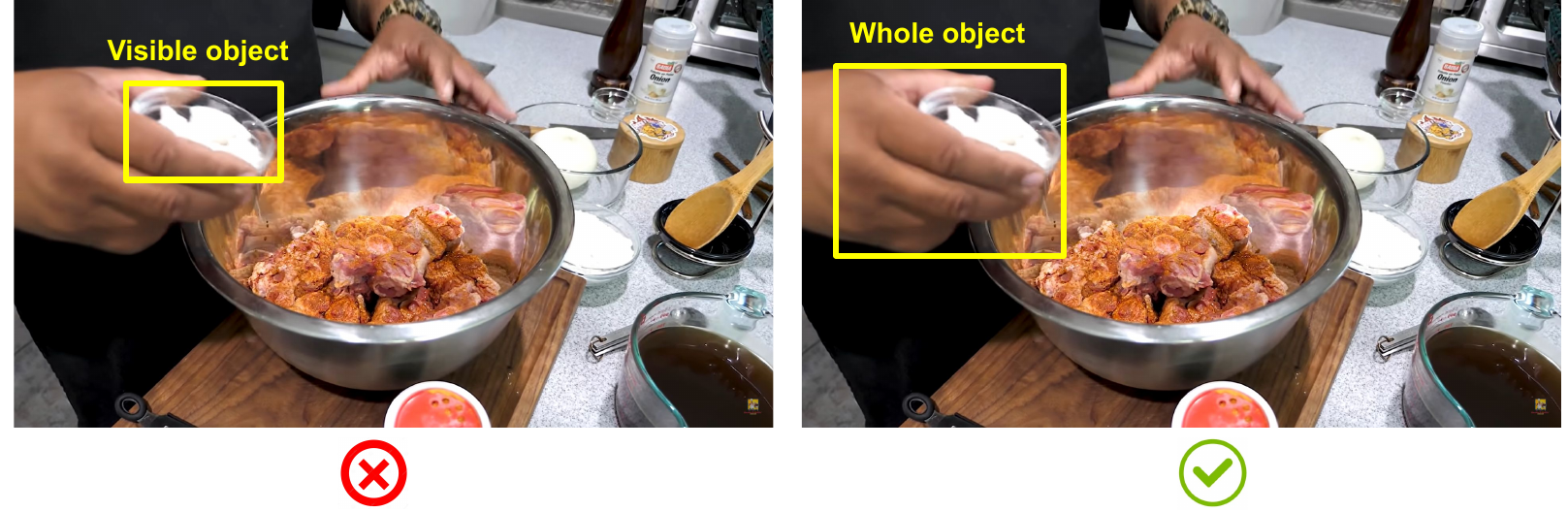}
    \caption{
    Visual example of how to annotate a bounding box when only part of the object is visible. (Video credit: Smokin' \& Grillin' wit AB~\cite{smokingrillin})
    }
    \label{fig:bb_guideline_1}
\end{figure}

Given the set of relevant frames that have been filtered by the first stage, Amazon Mechanical Turk workers are tasked to draw bounding boxes around the relevant regions given a natural language sentence. A visualization of the annotation interface can be found in Figure \ref{fig:annotation_interface} (this supplemental). The general guiding principles are listed as follows:
\begin{enumerate}
    \item Bounding boxes should only target the objects involved in the action as well as the entire hand / hands interacting with them. If only one hand is interacting with the object, the free hand should NOT be included.
    \item The bounding box should only enclose the object(s) being interacted with, not all objects mentioned in the sentence.
    \item It should be as tight as possible.
    \item In cases where the specified object is partially hidden, the bounding box should still enclose the entire hand and the object. For example in Figure \ref{fig:bb_guideline_1} (this supplemental), some parts of the object are occluded by her hand. However, the bounding box still includes the entire hand.
\end{enumerate}

Note that the HowTo100M dataset contains videos that overlap with those of YouCook2. To maintain a proper evaluation setting, overlapping videos are removed from the HowTo100M dataset. In our dataset, an interaction is defined as an action that involves the manipulation of at least one object. 

\subsection{Dataset statistics} 
\label{data_stats}
We show dataset statistics in Table \ref{table:dataset-statistics}. In general, our dataset contains approximately 256 video segments that are split among 92 videos. Each video segment is denoted by its start and end times and corresponds to a single narration. The vocabulary encompasses 54 unique verbs and 73 object categories. Since the vocabulary is extracted from the YouCook2 dataset descriptions, they are mainly restricted to the culinary domain.

\begin{table}[t!]
\centering
 \begin{tabular}{| c | c |} 
 \hline
 \# segments & 256 \\
 \hline
 \# videos & 92 \\
 \hline
 Average segment length & 24 sec. \\
 \hline
 \# frames & 6238 \\
 \hline
 \# verbs & 54 \\
 \hline
 \# nouns & 73  \\
 \hline
\end{tabular}
\caption{
{\bf YouCook2-Interactions dataset statistics.} Notice that our collected evaluation dataset with bounding box annotations supports a range of described actions and objects.
}
\label{table:dataset-statistics}
\end{table}

\end{document}